\documentclass{article}

\usepackage{microtype}
\usepackage{graphicx}
\usepackage{subcaption}
\usepackage{booktabs} 

\usepackage{hyperref}

\usepackage[preprint]{icml2026}

\usepackage{amsmath}
\usepackage{amssymb}

\usepackage{amsfonts}
\usepackage{algorithm}
\usepackage{algorithmic}
\usepackage{array}
\usepackage{textcomp}
\usepackage{url}
\usepackage{verbatim}
\usepackage{multirow}
\usepackage{tabularx}
\usepackage{placeins}
\usepackage[table]{xcolor}

\usepackage{mathtools}
\usepackage{amsthm}

\usepackage[capitalize,noabbrev]{cleveref}

\theoremstyle{plain}

\theoremstyle{definition}

\theoremstyle{remark}

\icmltitlerunning{Auto-Configured Networks for Multi-Scale Multi-Output Time-Series
Forecasting}

\begin{document}

\twocolumn[
  \icmltitle{Auto-Configured Networks for Multi-Scale Multi-Output Time-Series Forecasting}



 \begin{icmlauthorlist}
  \icmlauthor{Yumeng Zhao}{neu,dmu}
  \icmlauthor{Shengxiang Yang}{dmu}
  \icmlauthor{Xianpeng Wang}{neu-lab}
\end{icmlauthorlist}

\icmlaffiliation{neu}{National Frontiers Science Center for Industrial Intelligence and Systems Optimization, Northeastern University, Shenyang, China}
\icmlaffiliation{dmu}{Innovative Artificial Intelligence Cluster, Digital Future Institute, De Montfort University, Leicester, U.K.}
\icmlaffiliation{neu-lab}{Key Laboratory of Data Analytics and Optimization for Smart Industry (Northeastern University), Ministry of Education, Shenyang, China}

\icmlcorrespondingauthor{Shengxiang Yang}{syang@dmu.ac.uk}
\icmlcorrespondingauthor{Xianpeng Wang}{wangxianpeng@ise.neu.edu.cn}

\icmlkeywords{nonstationary multivariate time series, distribution drift, dynamic learning, evolutionary optimization, sintering process modeling}

\vskip 0.3in
]



\printAffiliationsAndNotice{}  

\begin{abstract}

Industrial forecasting often involves multi-source asynchronous signals and multi-output targets, while deployment requires explicit trade-offs between prediction error and model complexity. 
Current practices typically fix alignment strategies or network designs, making it difficult to systematically co-design preprocessing, architecture, and hyperparameters in budget-limited training-based evaluations.
To address this issue, we propose an auto-configuration framework that outputs a deployable Pareto set of forecasting models balancing error and complexity. 
At the model level, a Multi-Scale Bi-Branch Convolutional Neural Network (MS--BCNN) is developed, where short- and long-kernel branches capture local fluctuations and long-term trends, respectively, for multi-output regression.
At the search level, we unify alignment operators, architectural choices, and training hyperparameters into a hierarchical-conditional mixed configuration space, and apply Player-based Hybrid Multi-Objective Evolutionary Algorithm (PHMOEA) to approximate the error--complexity Pareto frontier within a limited computational budget.
Experiments on hierarchical synthetic benchmarks and a real-world sintering dataset demonstrate that our framework outperforms competitive baselines under the same budget and offers flexible deployment choices.
\end{abstract}

\section{Introduction}
Multivariate time-series forecasting has been widely adopted in many domains, including industry, energy, weather, and finance, for state monitoring and trend prediction, and continues to benefit from advances in deep temporal modeling~\cite{pmlr2024Unsupervised}.
In industrial processes, predicting quality variables that are difficult to measure online from process variables is a key approach for process monitoring and quality prediction~\cite{SHARDT202311768}.
Existing deep forecasting methods span a broad range of architectures, including CNNs, RNNs, and Transformers, yet multi-source heterogeneity, multi-scale variations, and multi-output targets in real-world settings remain persistent challenges~\cite{LimZohren2021TSFDeepLearningSurvey}.
For instance, in industrial settings, different sensors are often sampled at different rates, which makes mapping them onto a unified temporal resolution an unavoidable step before modeling in practical systems~\cite{Kong2025DLTSFSurvey}.

To tackle this, a common engineering practice is to adopt a fixed strategy to align multi-scale signals onto a unified time grid, and then apply standard models for learning and prediction~\cite{HUANG2021Amulti-rate}.
However, different alignment strategies differ in how they preserve high-frequency fluctuations, suppress noise, and handle missingness, and their performance depends strongly on sampling sparsity, noise level, and cross-scale coupling, making it difficult to reliably select the most suitable preprocessing scheme across datasets or operating conditions based on manual experience alone~\cite{Liu2025Rethinking}.
Moreover, when jointly considering multi-scale representation structures, training hyperparameters, and deployment constraints, preprocessing choices interact with network structure and training, leading to a surge of candidate configurations whose training-based evaluations are costly~\cite{LU2025ASurvey}.
From an optimization perspective, this amounts to a budgeted multi-objective black-box optimization problem with mixed variables and feasibility constraints, where only a few evaluations are affordable to approach favorable error--complexity trade-offs.
Consequently, the goal of auto-configuration should go beyond tuning within a fixed network, and instead systematically select combinations of preprocessing--structure--hyperparameters in a unified search space, while providing diverse error--complexity trade-offs in the form of a Pareto model set to meet heterogeneous deployment needs.

However, mainstream automated machine learning settings are mostly developed around standardized search spaces and generic tasks~\cite{HE2021AutoML}, whereas industrial multi-scale forecasting often also involves preprocessing operators such as resampling or aggregation, making these paradigms difficult to adopt directly.
Meanwhile, evaluating each candidate configuration typically requires an almost full training and validation run, leaving a strictly limited computational budget and thereby calling for auto-configuration methods that can find high-quality error--complexity trade-offs with fewer evaluations.
Therefore, we focus on a multi-scale multi-output forecasting problem distilled from real-world sintering quality prediction, and proposes a multi-objective auto-configuration framework that aims to output a deployable Pareto model set under a limited computational budget.

This problem remains largely underexplored in the literature due to the following challenges:
\begin{itemize}
  \item In mixed spaces with discrete and continuous variables and hierarchical conditional dependencies, it remains challenging to build a comparable unified encoding and feasibility-preserving variation mechanisms; moreover, the absence of standardized benchmarks makes reproducible evaluation harder.
  \item Single-branch modeling often fails to capture short-term perturbations and long-term trends jointly, and weak input--output correlations in industrial data further hinder multi-output representation learning.
  \item It is also challenging to stably output a competitive error--complexity Pareto model set with fewer evaluations while avoiding search degeneration and local optima under a limited computational budget.
\end{itemize}

We propose an auto-configured forecasting framework that searches a configurable multi-scale forecasting network space via multi-objective evolutionary search to obtain an error--complexity Pareto model set.
Specifically, MS--BCNN uses two convolutional branches with short and long kernels to model local fluctuations and long-term trends for multi-output regression, and we unify preprocessing operators, architectural hyperparameters, branch fusion strategies, and training hyperparameters into a mixed configuration space with conditional constraints.
Finally, under a limited computational budget, we apply PHMOEA to stably approximate the error--complexity Pareto frontier, and validate the resulting Pareto model set on hierarchical synthetic benchmarks and real sintering data.

The main contributions of our work are summarized as follows:
\begin{itemize}
  \item We distill an industrial forecasting task with multi-scale inputs and multi-output targets from real-world sintering quality prediction, and formalize it as a budgeted multi-objective auto-configuration problem with a deployment-oriented Pareto model set as output.
  \item We propose an auto-configuration framework that unifies MS--BCNN's multi-scale representations with PHMOEA's budgeted multi-objective search in a single pipeline; extensive experiments on hierarchical synthetic benchmarks and real sintering data demonstrate its advantages under a limited computational budget.
  \item To reproducibly evaluate search behaviors in hierarchical mixed decision spaces, we construct hierarchical synthetic benchmark problems to systematically examine hierarchical conditional encoding and multi-objective search.
\end{itemize}

\section{Related Work}
\label{Section2}

Recent advances in multivariate time-series forecasting often combine decomposition modeling with efficient attention or Transformers to capture long-term trends and cross-series dependencies for more stable and accurate long-horizon prediction~\cite{zhou22g}.
Another common strategy splits long sequences into fixed-length temporal segments and models segments as units, reducing computation while strengthening the representation of local patterns~\cite{nie23patchtst}.
However, most studies still report a single operating point under fixed preprocessing and fixed architectures, and they rarely expose a deployable set of trade-offs between forecasting error and model complexity.

Multi-source asynchronous data can be viewed as irregular multivariate time series, where variables are sampled at different rates and their timestamps are often not aligned, which complicates modeling both temporal relationships and cross-variable dependencies~\cite{ansari23a,zhang24bw,li25bl}.
To handle such irregularity, a class of continuous-time models advances computation using the actual time gaps and updates internal representations when new data arrive, avoiding the need to interpolate or pad data onto a fixed time grid~\cite{ansari23a}.
When timestamps are misaligned across variables, recent methods further use graph- or hypergraph-based message passing to learn asynchronous correlations, reducing reliance on strict alignment and padding~\cite{zhang24bw,li25bl}.

For auto-configuration, hyperparameter optimization and architecture search often require repeatedly training and evaluating candidate configurations, which is expensive, calling for efficient budget-aware search and explicit modeling of multi-objective trade-offs~\cite{rakotoarison24a,lu24evoxbench}.
Recent work develops multi-objective optimization strategies, including multi-objective Bayesian optimization and scalarization, to better approximate Pareto frontiers and characterize trade-offs among objectives~\cite{park24k,lin24y}.
In addition, freeze--thaw style approaches improve search efficiency under expensive evaluations by allocating training resources progressively, offering useful resource-allocation principles for budget-constrained configuration search~\cite{rakotoarison24a}.
Building on these lines, we unify alignment and resampling, a multi-scale backbone, and training choices into a hierarchical conditional configuration space and conduct budgeted multi-objective evolutionary search to output a deployment-oriented Pareto model set that explicitly presents the error--complexity trade-off.

\section{Industrial Background and Problem Formulation}
\label{Section3}
Sintering is a key upstream step in ironmaking that converts blended fines into sinter for downstream use, and its coupled heat transfer, combustion, and gas--solid flow lead to strongly nonlinear and multi-scale dynamics~\cite{Xue2024Phase-Field,Olevsky2006Multi-Scale}.
In practice, sensors are sampled asynchronously at multiple rates, while terminal quality is obtained from low-frequency and delayed assays, motivating data-driven forecasting for monitoring and control.
Grounded in this industrial task, we distill a general problem of auto-configuring networks for multi-scale multi-output forecasting on multi-source asynchronous time series.

We denote the multi-source input as a collection of time series sampled at different rates:
\begin{equation}
\mathbf{X}=
\big[\mathbf{X}^{(1)},\mathbf{X}^{(2)},\ldots,\mathbf{X}^{(S)}\big],
\end{equation}
where $S$ is the number of sources and $\mathbf{X}^{(s)}\in\mathbb{R}^{T_s\times d_s}$ denotes the raw sequence of source $s$ with length $T_s$ and channel dimension $d_s$.
We map each source to a fixed length $L_p$ and concatenate channels to obtain $\tilde{\mathbf{X}}\in\mathbb{R}^{L_p\times D}$ with $D=\sum_{s=1}^S d_s$.
Let $\mathbf{y}\in\mathbb{R}^{K}$ denote the $K$-dimensional target vector for the sample, where $K$ is the number of target variables to be predicted jointly.
Given $\tilde{\mathbf{X}}$, a model $\mathcal{M}(\cdot;\,x)$ specified by configuration $x$ outputs:
\begin{equation}
\hat{\mathbf{y}}(x)=\mathcal{M}(\tilde{\mathbf{X}};\,x)\in\mathbb{R}^{K}.
\label{eq:output-y-sec2}
\end{equation}

Furthermore, the configuration process exhibits conditional activation: parameters become feasible only when their associated operators are selected.
This induces a hierarchical mixed search space $\mathcal{X}$, whose formal definition is given in Appendix~\ref{app:hier-decision}.
Meanwhile, reducing forecasting error in real deployments often increases model complexity and inference cost, while lightweight models reduce cost but may sacrifice accuracy, so the objectives conflict and form a canonical multi-objective trade-off.
Accordingly, we formalize auto-configuration as a bi-objective optimization problem over a hierarchical conditional mixed space:
\begin{equation}
\min_{x\in\mathcal{X}} \, F(x) = \big( f_1(x),\, f_2(x) \big).
\label{eq:mo-problem-sec2}
\end{equation}
Here, $f_1(x)$ is the validation mean squared error (MSE) and $f_2(x)$ is the number of trainable parameters.
\begin{equation}
f_1(x)=\mathrm{MSE}(x)
=\frac{1}{N_{\mathcal{D}}K}\sum_{n=1}^{N_{\mathcal{D}}}\left\lVert \mathbf{y}_n-\hat{\mathbf{y}}_n(x)\right\rVert_2^2,
\label{eq:f1-mse-sec2}
\end{equation}
\begin{equation}
f_2(x)=P(x)=\sum_{\theta\in \Theta(x)} \mathrm{numel}(\theta).
\label{eq:f2-param-sec2}
\end{equation}
Here, $N_{\mathcal{D}}$ is the number of validation samples and $\Theta(x)$ is the set of trainable parameter tensors decoded from $x$; $\mathrm{numel}(\theta)$ is the number of scalar entries in $\theta$ (Appendix~\ref{app:param-count}).
We apply per-generation normalization only for search and it does not change Pareto dominance (Appendix~\ref{app:norm-objectives}).

\section{Method}
\label{Section4}

\subsection{Framework Overview}
\label{sec:overview}

\begin{figure*}[t]
  \centering
  \includegraphics[width=\linewidth]{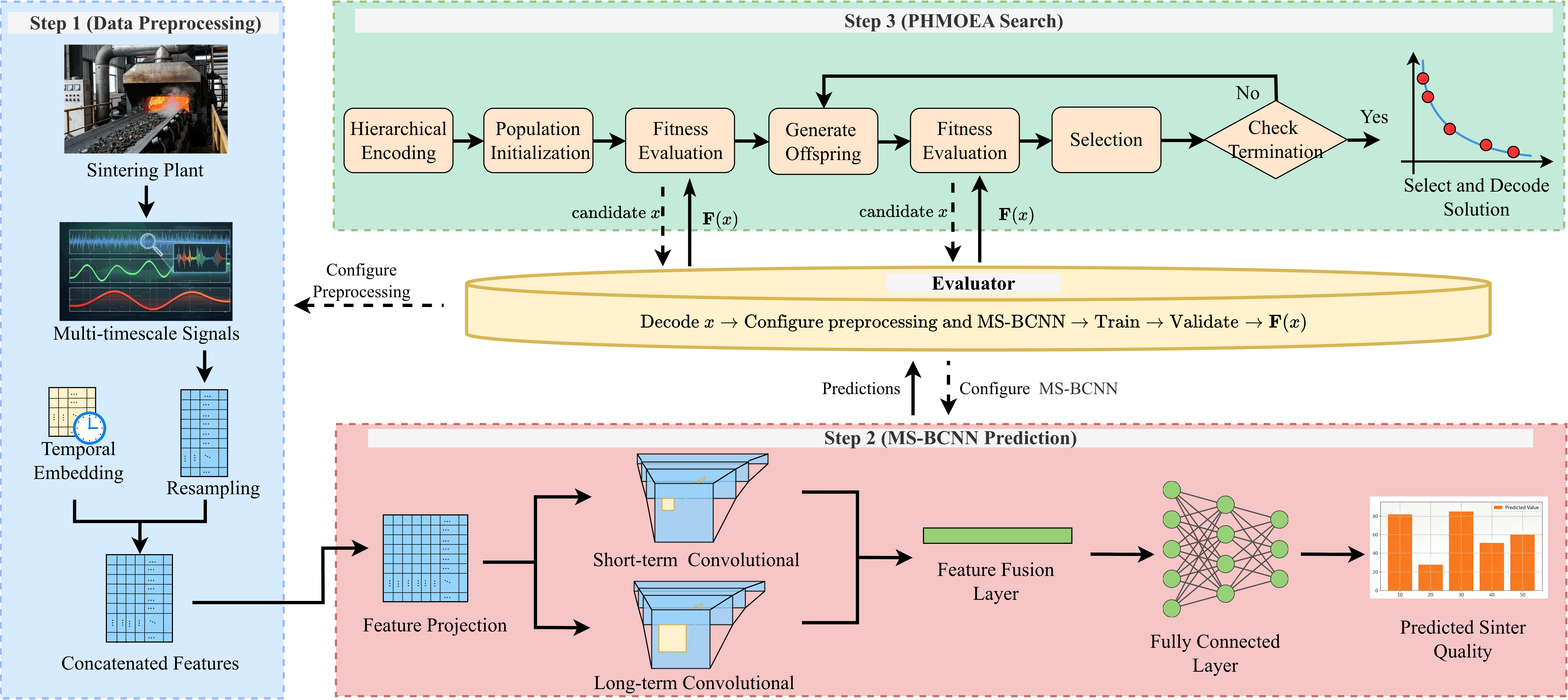}
  \caption{Evaluator-driven auto-configuration framework for multi-scale multi-output time-series forecasting.
PHMOEA proposes a configuration $x$, and the evaluator decodes $x$ to configure preprocessing, MS--BCNN, and training, returning $F(x)$ for selection and iteration.
Solid arrows denote data or prediction flows, and dashed arrows denote configuration flows.}
  \label{fig:framework}
\end{figure*}

As shown in Fig.~\ref{fig:framework}, we unify candidate proposal, evaluator-based evaluation, election and update into a closed loop for auto-configuring multi-scale multi-output time-series forecasting models.
Given a training dataset $\mathcal{D}$, PHMOEA iteratively proposes candidate configurations $x$ in a hierarchical conditional mixed space; $x$ jointly specifies preprocessing, the MS--BCNN architecture, and training hyperparameters.
The evaluator decodes $x$ into an executable pipeline and performs one train--validate run, returning the bi-objective value $F(x)=(f_1(x),f_2(x))$ for selection and update, thereby approaching the error--complexity Pareto front.
Under the joint constraints of a maximum number of generations and early stopping, the loop outputs a deployable Pareto set of models.

\subsection{Configurable MS--BCNN}
\label{sec:msbcnn}

This subsection introduces MS--BCNN, a configurable multi-scale convolutional forecasting backbone searched and evaluated by PHMOEA offline.

On the input side, we resample and align the raw sequence of each data source to a unified length $L_p$ using a configurable alignment operator, so that subsequent convolutions operate on a consistent time index.
The alignment operator and its parameters are included in the configuration space as discrete choices.
Beyond process signals, we concatenate standard sine--cosine periodic embeddings with the input features to encode sample-level time information.
We concatenate the aligned sequences from all sources along the feature dimension to form $\tilde{\mathbf{X}}$, and apply a per-timestep linear projection to map channels to $C_0$ before feeding the result into the convolutional backbone.

In the backbone, each layer uses parallel short-kernel and long-kernel convolutions to capture fast local fluctuations and slow global trends, with configurable kernel sizes, normalization, and activations.
After multiple bi-branch layers, we fuse the final representations $h^{\text{short}}$ and $h^{\text{long}}$ into $h$ using a configurable fusion operator:
\begin{equation}
h=
\mathrm{Fuse}_{\mathrm{conv}}\!\left(
h^{\text{short}},\, h^{\text{long}}
\right).
\label{eq:fuse}
\end{equation}

Finally, we flatten $h$ into a feature vector $f$ and use a lightweight prediction head to output a $K$-dimensional prediction vector.
In a single configuration space, we jointly search input alignment, architectural options such as kernel sizes, normalization, and activations, as well as training hyperparameters including the loss function and the learning-rate schedule.
To control the space size, we fix the number of layers and the stacking order, and provide full specifications and instantiation details in Appendix~\ref{app:search-space}.

\subsection{Hierarchical Conditional Configuration Space}
\label{sec:space}

\noindent\textbf{Hierarchy and conditional dependencies.}

We organize configuration variables by levels $\mathcal{L}$, and denote level-$\ell$ variables by $x^{(\ell)}$.
A conditional variable becomes active only when its parent discrete choice is selected; we formalize this with a binary activity indicator $a_{j}^{(\ell)}(x)\in\{0,1\}$.
During decoding, we mask all variables with $a_{j}^{(\ell)}(x)=0$ so that every decoded configuration is valid.
The complete variable list and dependency graph are provided in Appendix Table~\ref{tab:decision-space}.
For a unified notation, the $i$-th individual is represented as a hierarchical decision vector:
\begin{equation}
x_i = \big( x_{i,j}^{(\ell)} \big)_{\ell \in \mathcal{L},\, j = 1,\dots,n_\ell},
\label{eq:hierarchical}
\end{equation}
where $n_\ell$ is the number of variables at level $\ell$, and $x_{i,j}^{(\ell)}$ encodes a continuous, discrete, or conditional decision.

\noindent\textbf{Encoding and decoding.}

To apply the same evolutionary operators to discrete, continuous, and conditional variables, we adopt a unified discrete index encoding and decode indices back to executable configurations.
Discrete variables are encoded by candidate indices; each continuous variable is discretized into $K_d$ bins (linear- or log-scale as appropriate) and encoded by the bin index, which is mapped back to a concrete value during decoding.
For conditional variables, decoding activates them only when their parent discrete choices are active; otherwise, they are masked to keep every decoded configuration valid.
Implementation details of the discretization and the linear and log mappings are provided in Appendix~\ref{app:encdec}.

\noindent\textbf{Hierarchical repair.}

To ensure executability, we perform hierarchical repair around decoding and use Lamarckian write-back~\cite{elsken2019efficient}, where repaired values are written back to the individual representation and inherited in subsequent evolution.
For continuous variables that are sensitive to training, we keep an empirically determined safe range and clip out-of-range values to avoid numerical instability or obviously invalid evaluations.
For inactive conditional variables, we apply semantic freezing by marking them invalid to skip computation; once re-activated, we restore their most recent valid values to preserve genetic continuity.

\noindent\textbf{Deduplication.}

To reduce duplicate evaluations, we canonicalize each repaired-and-decoded executable configuration and compute a hash key~\cite{carter1979universal} over its deterministic serialization; candidates whose keys have appeared before are rejected before training.
We deduplicate at the executable-configuration level, so different genotypes that decode to the same configuration are evaluated only once.
If a candidate is infeasible or duplicated, we retry offspring generation up to $n_{\text{trial}}$ attempts; if all attempts fail, we skip this offspring slot to avoid stalling (Appendix~\ref{app:dedup}).

\noindent\textbf{Adaptive refinement for continuous variables.}

To enable coarse-to-fine search, we adaptively refine the discretization of each continuous variable by maintaining interval partitions and refining them based on how non-dominated solutions concentrate over intervals.
Specifically, we compute the fraction $f^{(\ell)}_{d,k}$ of the current generation's non-dominated solutions falling into the $k$-th interval; when an interval satisfies $f^{(\ell)}_{d,k}>\delta_h$ for $H$ consecutive generations, we trigger further refinement on that interval.
We maintain a cross-generation state $\mathcal{S}$ to implement the $H$-generation persistence and avoid spurious triggers due to stochastic fluctuations; implementation details are provided in Appendix~\ref{app:refine}.
Representative values follow the same linear and log discretization in Appendix~\ref{app:encdec}, enabling progressively finer search under the unified encoding.

\subsection{PHMOEA: Framework and Complexity}
\label{sec:phmoea}

\noindent\textbf{Player scoring standards.}

To convert multi-objective performance into archive-update weights, we design a stage-dependent score $R_i$, which updates the cross-generation player archives $(\mathcal{H},\mathcal{C})$ and guides stratified offspring sampling.
The stage factor is $\phi_t=t/T_{\max}$, controlling the stage-wise weighting.
We further use within-generation normalized objectives $\hat{f}_m(x_i)$ and a normalized crowding measure $\hat{c}_i$ (Appendix~\ref{app:norm-objectives}).

We define a rank--diversity score $s_i$ and an objective-guided score $g_i$, and blend them across stages to obtain $R_i$:
\begin{align*}
s_i &= \frac{1}{1+r_i} + \lambda\,\hat{c}_i,\\
g_i &= w\,\hat{f}_1(x_i) + (1-w)\,\hat{f}_2(x_i),
\end{align*}
\begin{equation}
R_i=
\begin{cases}
s_i, & \phi_t < \kappa_1,\\[0.2em]
\alpha_t s_i + (1-\alpha_t) g_i, & \kappa_1 \le \phi_t < \kappa_2,\\[0.2em]
g_i + \gamma\,\hat{c}_i, & \phi_t \ge \kappa_2,
\end{cases}
\label{eq:score-ri}
\end{equation}
where $r_i$ is the non-dominated rank; $0\le \kappa_1<\kappa_2\le 1$ are stage-switching thresholds, and $\alpha_t=(\kappa_2-\phi_t)/(\kappa_2-\kappa_1)$.
Here $\lambda,w,\gamma$ are hyperparameters; we normalize $R_i$ over the population to obtain weights $\omega_i$ (with $\sum_i \omega_i=1$) for archive updates.

\noindent\textbf{Generate offspring with player tracking.}
\phantomsection\label{para:pt-gen}

PHMOEA treats each dimension-wise candidate $a_{j,k}\in A_j$ as a player and maintains two global player-keyed archives: a heat archive $\mathcal{H}$ and a count archive $\mathcal{C}$.
Given the parent population $P_t$ from environmental selection, we update $(\mathcal{H},\mathcal{C})$ using the individual weights $\{\omega_i\}$ by accumulating $\omega_i$ to the players appearing in $x_i$ and incrementing their occurrence counts.

For each dimension $j$, we partition candidates into a hot set $A_j^{\mathrm{hot}}$ (top-$q\%$ by heat), a cold set $A_j^{\mathrm{cold}}$ (bottom-$p\%$ by count), and a normal set $A_j^{\mathrm{norm}}$ (the remainder), and define the non-hot pool $A_j^{\mathrm{nh}}=A_j^{\mathrm{norm}}\cup A_j^{\mathrm{cold}}$.
Sampling from $A_j^{\mathrm{hot}}$ is uniform, whereas sampling from $A_j^{\mathrm{nh}}$ applies a cold-bonus factor $o$ and renormalizes within the pool:
\begin{equation}
P(a \mid A_j^{\mathrm{nh}})=\frac{\eta(a)}{\sum_{b\in A_j^{\mathrm{nh}}} \eta(b)},\quad
\eta(a)=\begin{cases}
o, & a\in A_j^{\mathrm{cold}},\\
1, & a\in A_j^{\mathrm{norm}}.
\end{cases}
\label{eq:nh-sampling}
\end{equation}

At generation $t$, we generate $N$ offspring $Q_t$ from three sources with stage ratios $(\rho_{\mathrm{par}},\rho_{\mathrm{hot}},\rho_{\mathrm{nh}})$ determined by $\phi_t$ (same stage split as Eq.~\ref{eq:score-ri}), where $\rho_{\mathrm{par}}+\rho_{\mathrm{hot}}+\rho_{\mathrm{nh}}=1$.
We create $N_{\mathrm{par}}=\lfloor \rho_{\mathrm{par}}N\rfloor$ parent-operator offspring, $N_{\mathrm{hot}}=\lfloor \rho_{\mathrm{hot}}N\rfloor$ Hot-type offspring, and set $N_{\mathrm{nh}}=N-N_{\mathrm{par}}-N_{\mathrm{hot}}$ for NonHot-type offspring.
The parent-operator offspring are produced from $P_t$ via SBX crossover and polynomial mutation, while Hot-type and NonHot-type offspring are assembled by sampling one candidate per dimension from $\{A_j^{\mathrm{hot}}\}$ or $\{A_j^{\mathrm{nh}}\}$, respectively.
For assembled offspring, we apply variable-wise cross-pool mutation: each dimension is mutated with probability $e$ (up to $m_{\max}$ mutated dimensions per individual) by resampling from the opposite pool $A_j^{\bar{s}}$ (with $\bar{s}$ opposite to the assembling pool), followed by hierarchical repair and deduplication (Sec.~\ref{sec:space}) to ensure feasibility and uniqueness.
Finally, we perform environmental selection on $P_t\cup Q_t$ to obtain $P_{t+1}$ with population size $N$; see Algorithm~\ref{alg:pt-offspring} for the full procedure.

\begin{algorithm}[t]
\caption{Generate offspring with player tracking}
\label{alg:pt-offspring}
\begin{algorithmic}
\STATE {\bfseries Input:} Parents $P_t$; population size $N$; candidate sets $\{A_j\}_{j=1}^J$; heat/count archives $(\mathcal{H},\mathcal{C})$; stage factor $\phi_t$; cross-pool mutation probability $e$; max mutated dims $m_{\max}$.
\STATE {\bfseries Output:} Offspring $Q_t$; updated archives $(\mathcal{H},\mathcal{C})$.
\end{algorithmic}
\begin{algorithmic}[1]
\STATE Update archives $(\mathcal{H},\mathcal{C})$ by Eqs.~(\ref{eq:score-ri}).
\FOR{$j=1$ \textbf{to} $J$}
  \STATE Partition $A_j$ into $A_j^{\mathrm{hot}}, A_j^{\mathrm{norm}}, A_j^{\mathrm{cold}}$ using archives $(\mathcal{H},\mathcal{C})$;
  \STATE $A_j^{\mathrm{nh}} \leftarrow A_j^{\mathrm{norm}} \cup A_j^{\mathrm{cold}}$;
  \STATE Sample uniformly in $A_j^{\mathrm{hot}}$; sample from $A_j^{\mathrm{nh}}$ by Eq.~(\ref{eq:nh-sampling});
\ENDFOR
\STATE Compute counts $N_{\mathrm{par}}, N_{\mathrm{hot}}, N_{\mathrm{nh}}$ from $\phi_t$ and $N$;
\STATE $Q_t \leftarrow \emptyset$;
\STATE \textit{(Lines~10--12 apply hierarchical repair and deduplication to each new offspring.)}
\STATE $Q_t^{\mathrm{par}} \leftarrow \mathrm{Variation}(P_t, N_{\mathrm{par}})$ with SBX crossover and polynomial mutation;
\STATE $Q_t^{\mathrm{hot}} \leftarrow \mathrm{AssembleHot}(\{A_j^{\mathrm{hot}}\}_{j=1}^J, N_{\mathrm{hot}})$ with uniform sampling and cross-pool mutation $(e,m_{\max})$;
\STATE $Q_t^{\mathrm{nh}} \leftarrow \mathrm{AssembleNonHot}(\{A_j^{\mathrm{nh}}\}_{j=1}^J, N_{\mathrm{nh}})$ with sampling by Eq.~(\ref{eq:nh-sampling}) and cross-pool mutation $(e,m_{\max})$;
\STATE $Q_t \leftarrow Q_t^{\mathrm{par}} \cup Q_t^{\mathrm{hot}} \cup Q_t^{\mathrm{nh}}$;
\STATE \textbf{return} $Q_t$, $(\mathcal{H},\mathcal{C})$.
\end{algorithmic}
\end{algorithm}

\noindent\textbf{Main procedure.}

Algorithm~\ref{alg:phmoea-main} outlines PHMOEA.
We initialize and evaluate $P_0$, and initialize the archives $(\mathcal{H},\mathcal{C})$ and refinement state $\mathcal{S}$.
At generation $t$, we check the budgeted early-stopping criterion before offspring generation and stop if it is met (Appendix~\ref{app:early-stop}).
We update $\mathcal{S}$ and refine continuous variables when triggered, enabling coarse-to-fine search.
We generate $Q_t$ from $P_t$ via Algorithm~\ref{alg:pt-offspring} (updating $(\mathcal{H},\mathcal{C})$), evaluate $Q_t$, and select $P_{t+1}$ from $P_t\cup Q_t$.
Finally, we output the non-dominated set of the last population as $\mathcal{P}^\star$.
The Evaluator is the same as in Fig.~\ref{fig:framework}: each evaluation runs one train--validate, returns $F(x)$, and counts as one function evaluation.

\begin{algorithm}[t]
\caption{Main algorithm of PHMOEA}
\label{alg:phmoea-main}
\begin{algorithmic}
\STATE {\bfseries Input:} Dataset $\mathcal{D}$; candidate sets $\{A_j\}_{j=1}^J$; population size $N$; max generations $T_{\max}$; stage factor $\phi_t$; 
cross-pool mutation probability $e$; max mutated dims $m_{\max}$.
\STATE {\bfseries Output:} Pareto set $\mathcal{P}^\star$.
\end{algorithmic}
\begin{algorithmic}[1]
\STATE $P_0 \leftarrow \mathrm{InitRandomPop}(N,\{A_j\}_{j=1}^J)$;
\STATE $P_0 \leftarrow \mathrm{Evaluator}(P_0,\mathcal{D})$;
\STATE $((\mathcal{H},\mathcal{C}),\mathcal{S}) \leftarrow \mathrm{InitArchives}(\{A_j\}_{j=1}^J)$;
\FOR{$t=0$ \textbf{to} $T_{\max}-1$}
    \IF{$\mathrm{EarlyStop}(P_t)$}
        \STATE \textbf{break};
    \ENDIF
    \STATE $\mathcal{S} \leftarrow \mathrm{UpdateRefineState}(\mathcal{S}, P_t)$;
    \IF{$\mathrm{RefineCheck}(P_t,\mathcal{S})$}
        \STATE $\{A_j\}_{j=1}^J \leftarrow \mathrm{AdaptiveRefine}(\{A_j\}_{j=1}^J, P_t, \mathcal{S})$;
    \ENDIF
    \STATE \parbox[t]{0.95\linewidth}{%
    $(Q_t,(\mathcal{H},\mathcal{C})) \leftarrow \mathrm{GenerateOffspring}(P_t, N,$\\
    $\{A_j\}_{j=1}^J, (\mathcal{H},\mathcal{C}), \phi_t, e, m_{\max})$ by Algorithm~\ref{alg:pt-offspring};}
    \STATE $Q_t \leftarrow \mathrm{Evaluator}(Q_t,\mathcal{D})$;
    \STATE $P_{t+1} \leftarrow \mathrm{EnvironmentalSelect}(P_t \cup Q_t, N)$;
\ENDFOR
\STATE $\mathcal{P}^\star \leftarrow \mathrm{ExtractPF}(P)$;
\STATE \textbf{return} $\mathcal{P}^\star$.
\end{algorithmic}
\end{algorithm}

\noindent\textbf{Complexity.}

The runtime is dominated by training-based function evaluations: letting $C$ be the average cost of one train--validate run, the overall time is $O(\mathrm{FEs}\cdot C)$.
In the worst case, $\mathrm{FEs}=T_{\max}N$, yielding the standard bound $O(T_{\max}NC)$; in practice, repair, deduplication, and early stopping reduce the number of candidates admitted to evaluation, thereby lowering effective evaluations.
The remaining genetic and archive-maintenance overhead is lightweight relative to evaluation and does not change the dominant term (Appendix~\ref{app:complexity}).

\section{Experiments} 
\label{Section5}

\subsection{Experimental Setup}
\label{sec:exp-setup}

\textbf{Real-world dataset.}

We collect 2283 production-cycle samples from a sintering workshop in a large steel enterprise to evaluate multi-output quality forecasting in a real industrial setting.
Each sample is associated with five terminal quality targets (TFe, FeO, SiO$_2$, CaO, and basicity), whose assay-based ground-truth values become available with a delay after each production cycle.
We use a chronological split as the primary evaluation setting, and additionally report results on a randomly shuffled split as a complementary control under an approximately independent and identically distributed (i.i.d.) assumption.
Input variables are collected from on-site measurement systems, and the complete input--target list as well as preprocessing and split details are provided in Appendix~\ref{app:sinter-dataset}.

\textbf{Benchmarks.}

We evaluate hierarchical-conditional search on hierarchical synthetic benchmarks and the real sintering task.
To our knowledge, there are no widely used public benchmarks that explicitly capture hierarchical-conditional decision structures; thus, we construct hierarchical variants of classical DTLZ functions~\cite{deb2005dtlz} to analyze search behavior under controlled Pareto-front geometries.
Specifically, we use bi-objective H-DTLZ2 (smooth convex Pareto front) and H-DTLZ7 (disconnected and multimodal Pareto front) to cover complementary front shapes (definitions and the hierarchical projection $\Pi(\cdot)$ are given in Appendix~\ref{app:benchmarks-hdtlz}).
On the real sintering task, PHMOEA searches hierarchical conditional configurations spanning preprocessing operators, the MS--BCNN architecture, and training hyperparameters, and outputs a deployable Pareto model set.
Precise definitions of the resampling operators in the preprocessing module are provided in Appendix~\ref{app:resampling-ops}.

\textbf{Baselines, metrics, and availability.}
In evolutionary-search comparisons, we evaluate PHMOEA against four widely used multi-objective evolutionary algorithms, including MOEA/D~\cite{zhang2007moead}, NSGA-II~\cite{deb2002nsga2}, NSGA-III~\cite{deb2014nsga3}, and SMS-EMOA~\cite{beume2007smsemoa}.
We measure performance using inverted generational distance (IGD)~\cite{coello2004igd} and hypervolume (HV)~\cite{zitzler1999hypervolume}, and report function evaluations (FEs) as the search cost (formal definitions are provided in Appendix~\ref{appendix:evo_metrics}).
Since the true Pareto front is unavailable for the real sintering task, we compute IGD using an approximate reference front constructed by merging all solutions produced by all methods and then filtering the non-dominated (and duplicate-free) subset.
For forecasting-performance comparisons, we compare the MS--BCNN selected from PHMOEA's Pareto set against OB-ISSID~\cite{wang2025obissid}, Ventingformer~\cite{dai2025ventingformer}, Transformer~\cite{vaswani2017attention}, LSTM~\cite{hochreiter1997lstm}, and GRU-PLS~\cite{yang2022grupls}.
We evaluate forecasting performance using MSE, mean absolute error (MAE), mean absolute percentage error (MAPE),  and overall normalized metrics (NMSE and NMAE), with all metric definitions provided in Appendix~\ref{appendix:model_metrics}.
For reproducibility, we provide code, the normalized dataset, and preprocessing scripts in the supplementary material (detailed settings are provided in Appendix~\ref{app:exp-settings}).

\subsection{Search Quality and Efficiency}
\label{sec:exp-evo}

We compare PHMOEA against representative evolutionary baselines on both synthetic benchmarks and the real-world sintering task, and summarize the results in terms of search quality (IGD and HV) and evaluation efficiency (FEs).

\begin{figure}[t]
  \centering
  \includegraphics[width=\linewidth]{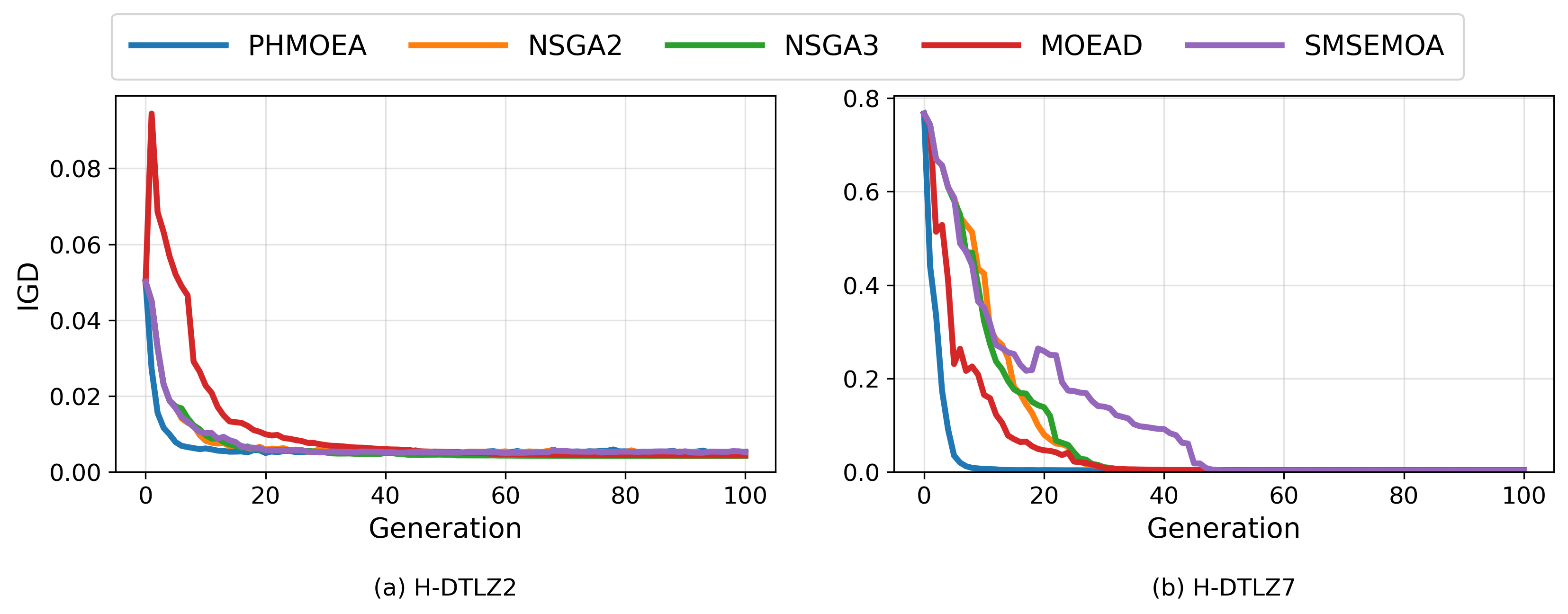}
  \caption{IGD curves along evolution on two H-DTLZ benchmark problems. }
  \label{fig:igd_curves_hdtlz_27}
\end{figure}

First, on the synthetic benchmarks, PHMOEA converges faster and stabilizes earlier at a low IGD level (Fig.~\ref{fig:igd_curves_hdtlz_27}).
At the final generation, PHMOEA achieves IGD and HV that are close to strong baselines, even though it is not the best on every metric (Appendix Table~\ref{tab:evo_hdtlz_27}).

Second, on the real-world sintering task, PHMOEA achieves the best or tied-best search quality and significantly outperforms several baselines (Table~\ref{tab:evo_cmp}).
Here $\downarrow$ and $\uparrow$ indicate lower and higher is better; $^{+}$, $^{-}$, $^{=}$ mean PHMOEA is significantly better, worse, or no different than the baseline.
Notably, MOEA/D remains markedly worse under the same evaluation budget, suggesting insufficient convergence within the allotted evaluations on this task.

Third, on the real-world sintering task, PHMOEA is substantially more evaluation-efficient under the same budget and stopping rule, which is particularly important when each evaluation requires a full train-and-validate run (Table~\ref{tab:evo_cmp}).
As a result, PHMOEA reaches a diminishing-returns region earlier and yields an equally strong or better trade-off set at lower evaluation cost.

\begin{table}[t]
  \centering
  \caption{Sintering-task results of multi-objective evolutionary algorithms (mean$\pm$std).}
  \label{tab:evo_cmp}
  \small
  \setlength{\tabcolsep}{3.5pt}
  \begin{tabular}{lccc}
    \toprule
    Algorithm & IGD$\downarrow$ & HV$\uparrow$ & FEs$\downarrow$ \\
    \midrule
    PHMOEA   & \cellcolor{gray!25}0.0163$\pm$0.0026 &
               \cellcolor{gray!25}0.986$\pm$0.005 &
               \cellcolor{gray!25}1113.6$\pm$1.9 \\
    MOEA/D   & 0.1400$\pm$0.0150$^{+}$ &
               0.781$\pm$0.020$^{+}$ &
               1500 \\
    NSGA-II  & 0.0308$\pm$0.0040$^{+}$ &
               0.984$\pm$0.006$^{=}$ &
               1500 \\
    NSGA-III & 0.0282$\pm$0.0035$^{+}$ &
               \cellcolor{gray!25}0.986$\pm$0.005$^{=}$ &
               1500 \\
    SMS-EMOA & 0.0175$\pm$0.0030$^{=}$ &
               0.980$\pm$0.006$^{=}$ &
               1500 \\
    \bottomrule
  \end{tabular}
\end{table}

\subsection{Forecasting Performance}
\label{sec:exp-forecast}

This section evaluates multi-target forecasting on the real-world sintering dataset under two settings: a shuffled split as an approximate i.i.d.\ reference and a chronological split that better reflects deployment.

First, under the shuffled setting, most baselines already achieve competitive overall errors, indicating that this reference comparison reflects modeling capacity rather than overly weak competitors (Table~\ref{tab:static_overall}).
In addition, MS--BCNN ranks first on the overall metrics, and the per-target breakdown and full visualizations of the same experiments are provided in the appendix (Appendix Table~\ref{tab:static_full} and Appendix Fig.~\ref{fig:static_pred_all}).

Second, under the chronological setting, all methods incur larger overall errors, reflecting the practical difficulty caused by non-stationarity and temporal shifts (Table~\ref{tab:dynamic_overall}).
Nevertheless, MS--BCNN remains the best on the composite normalized errors, while Ventingformer is slightly better on the relative-error metric, suggesting complementary strengths under different error characterizations (Table~\ref{tab:dynamic_overall}).

\begin{figure}[htbp]
\centering
\includegraphics[width=\linewidth]
{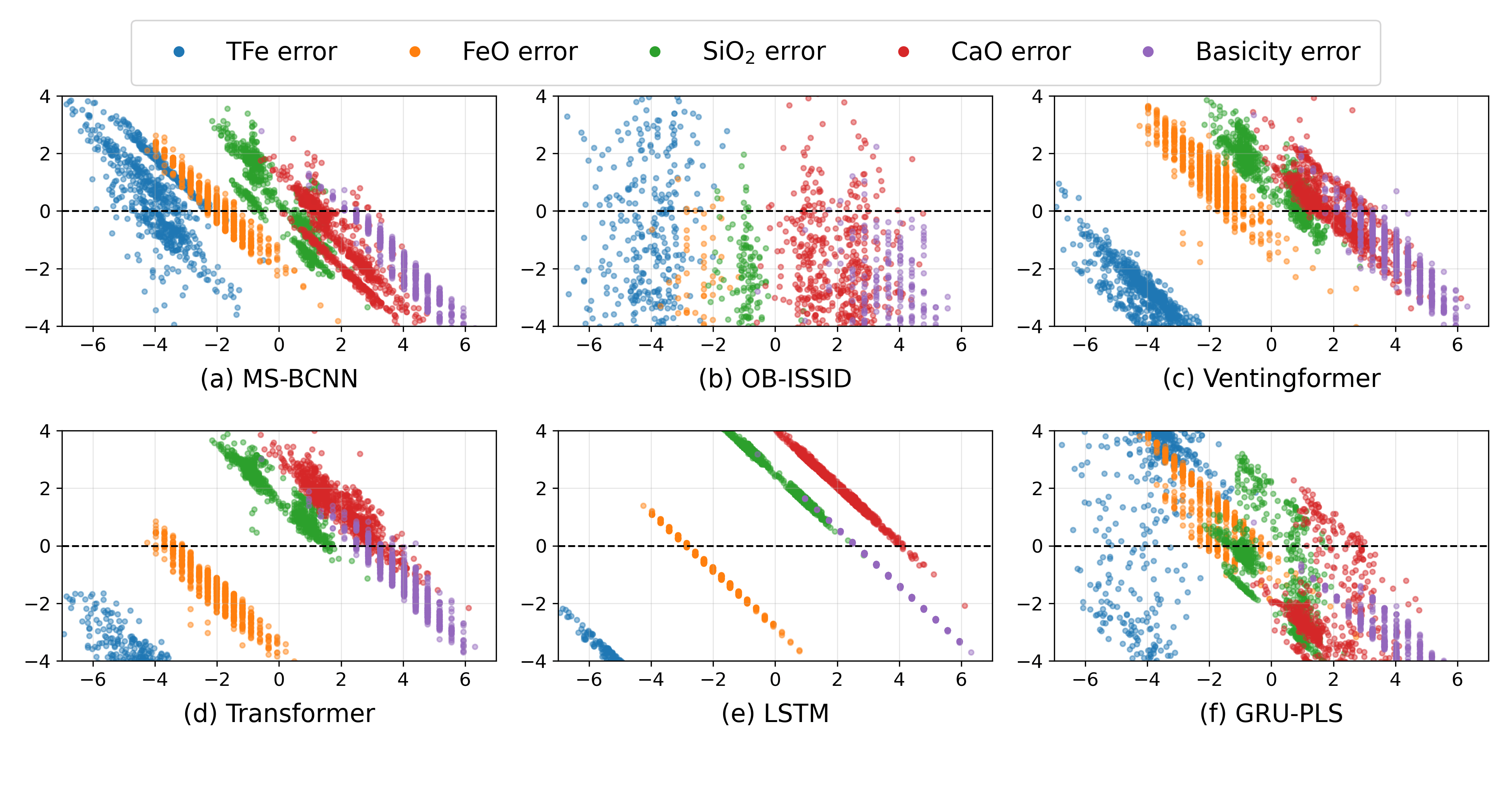}
\caption{Error scatter plot on the real-world sintering task under the chronological evaluation setting.}
\label{fig:dynamic_scatter}
\end{figure}

Third, the error scatter plot complements the numerical results by revealing error patterns: MS--BCNN exhibits a more concentrated distribution closer to the zero-error region, whereas several baselines show more dispersed or structured deviations (Fig.~\ref{fig:dynamic_scatter}).
Detailed per-target numbers and full time-series plots for the same chronological evaluation are provided in the appendix (Appendix Table~\ref{tab:dynamic_full} and Appendix Fig.~\ref{fig:static_true_pred}).

Overall, the shuffled setting serves as an approximate i.i.d.\ reference that also confirms the strength of the baselines, while the chronological evaluation better reflects deployment difficulty; across both settings, MS--BCNN demonstrates more reliable overall accuracy and stability.

\begin{table}[t]
  \centering
  \caption{Overall forecasting performance on the real-world sintering dataset with randomly shuffled samples (mean$\pm$std).}
  \label{tab:static_overall}
  \small
  \setlength{\tabcolsep}{4.2pt}
  \renewcommand{\arraystretch}{1.05}
  \begin{tabular}{lccc}
    \toprule
    Model & NMSE$\downarrow$ & NMAE$\downarrow$ & MAPE(\%)$\downarrow$ \\
    \midrule
    MS-BCNN
      & \cellcolor{gray!50}{$0.268 \pm 0.217$}
      & \cellcolor{gray!50}{$0.348 \pm 0.138$}
      & \cellcolor{gray!50}{$1.19 \pm 0.04$} \\
    OB-ISSID
      & $0.304 \pm 0.201$
      & $0.381 \pm 0.135$
      & $1.31 \pm 0.07$ \\
    Ventingformer
      & $0.279 \pm 0.199$
      & $0.366 \pm 0.136$
      & $1.25 \pm 0.07$ \\
    Transformer
      & $0.456 \pm 0.177$
      & $0.483 \pm 0.103$
      & $1.65 \pm 0.17$ \\
    LSTM
      & $0.380 \pm 0.220$
      & $0.421 \pm 0.142$
      & $1.44 \pm 0.08$ \\
    GRU-PLS
      & $0.462 \pm 0.171$
      & $0.512 \pm 0.082$
      & $1.74 \pm 0.08$ \\
    \bottomrule
  \end{tabular}
\end{table}

\begin{table}[t]
  \centering
  \caption{Overall forecasting performance on the real-world sintering dataset under the chronological evaluation setting (mean$\pm$std).}
  \label{tab:dynamic_overall}
  \small
  \setlength{\tabcolsep}{4.2pt}
  \renewcommand{\arraystretch}{1.05}
  \begin{tabular}{lccc}
    \toprule
    Model & NMSE$\downarrow$ & NMAE$\downarrow$ & MAPE(\%)$\downarrow$ \\
    \midrule
    MS-BCNN
      & \cellcolor{gray!50}{$4.446 \pm 2.193$}
      & \cellcolor{gray!50}{$1.682 \pm 0.490$}
      & $3.02 \pm 0.66$ \\
    OB-ISSID
      & $119.7 \pm 101.7$
      & $8.003 \pm 3.659$
      & $13.57 \pm 7.50$ \\
    Ventingformer
      & $4.789 \pm 4.111$
      & $1.698 \pm 0.746$
      & \cellcolor{gray!50}{$2.89 \pm 0.87$} \\
    Transformer
      & $7.581 \pm 8.678$
      & $2.180 \pm 1.304$
      & $3.43 \pm 0.76$ \\
    LSTM
      & $114.8 \pm 206.4$
      & $6.334 \pm 7.545$
      & $4.23 \pm 1.53$ \\
    GRU-PLS
      & $10.714 \pm 6.280$
      & $2.795 \pm 1.053$
      & $4.61 \pm 0.18$ \\
    \bottomrule
  \end{tabular}
\end{table}

\subsection{Ablation Study}
\label{sec:ablation}

\begin{figure}[t]
  \centering

  \begin{minipage}{0.48\linewidth}
    \centering
    \includegraphics[width=\linewidth]{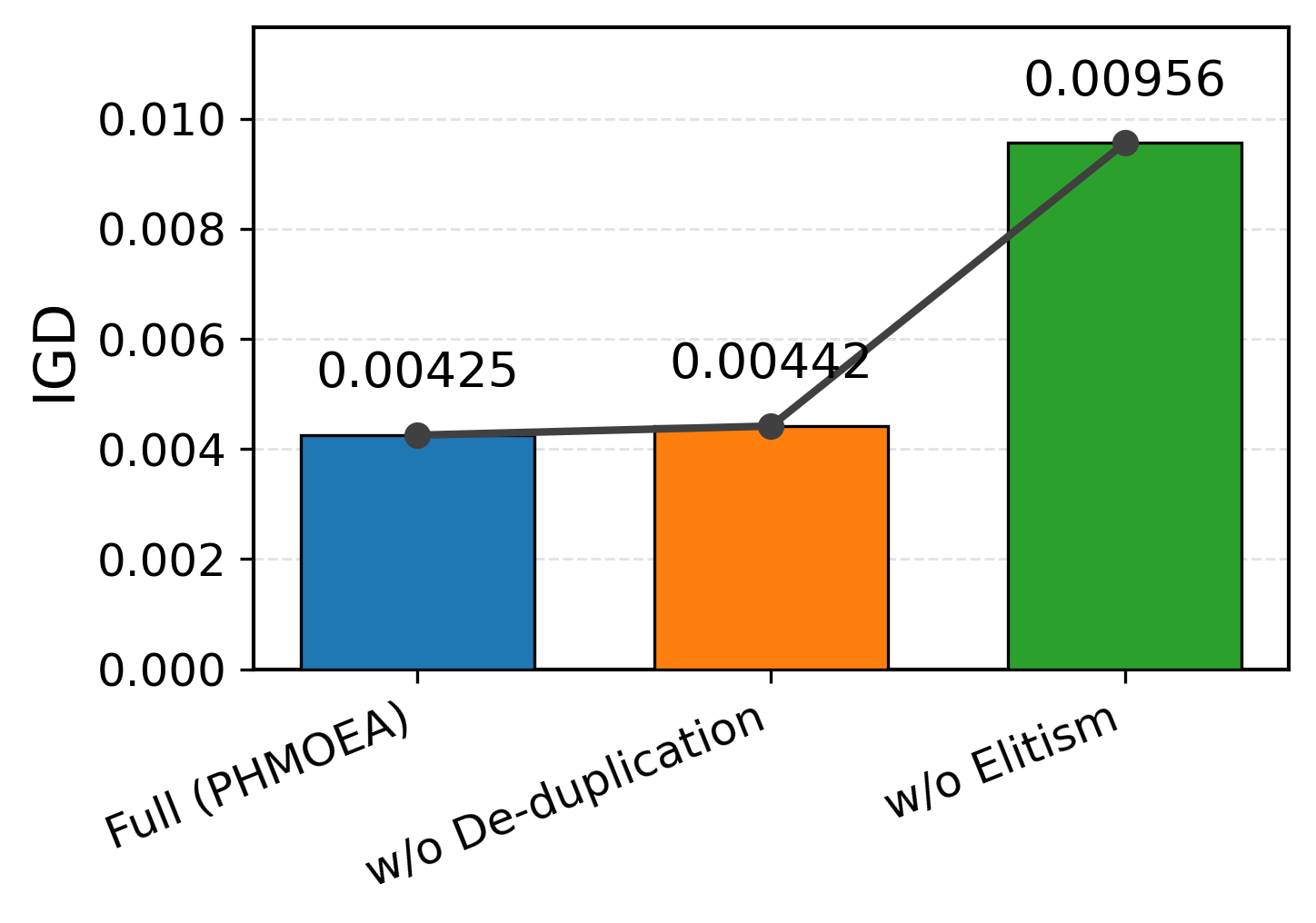}
    \\[0.2em]
    \small (a) PHMOEA ablation
  \end{minipage}
  \hfill
  \begin{minipage}{0.48\linewidth}
    \centering
    \includegraphics[width=\linewidth]{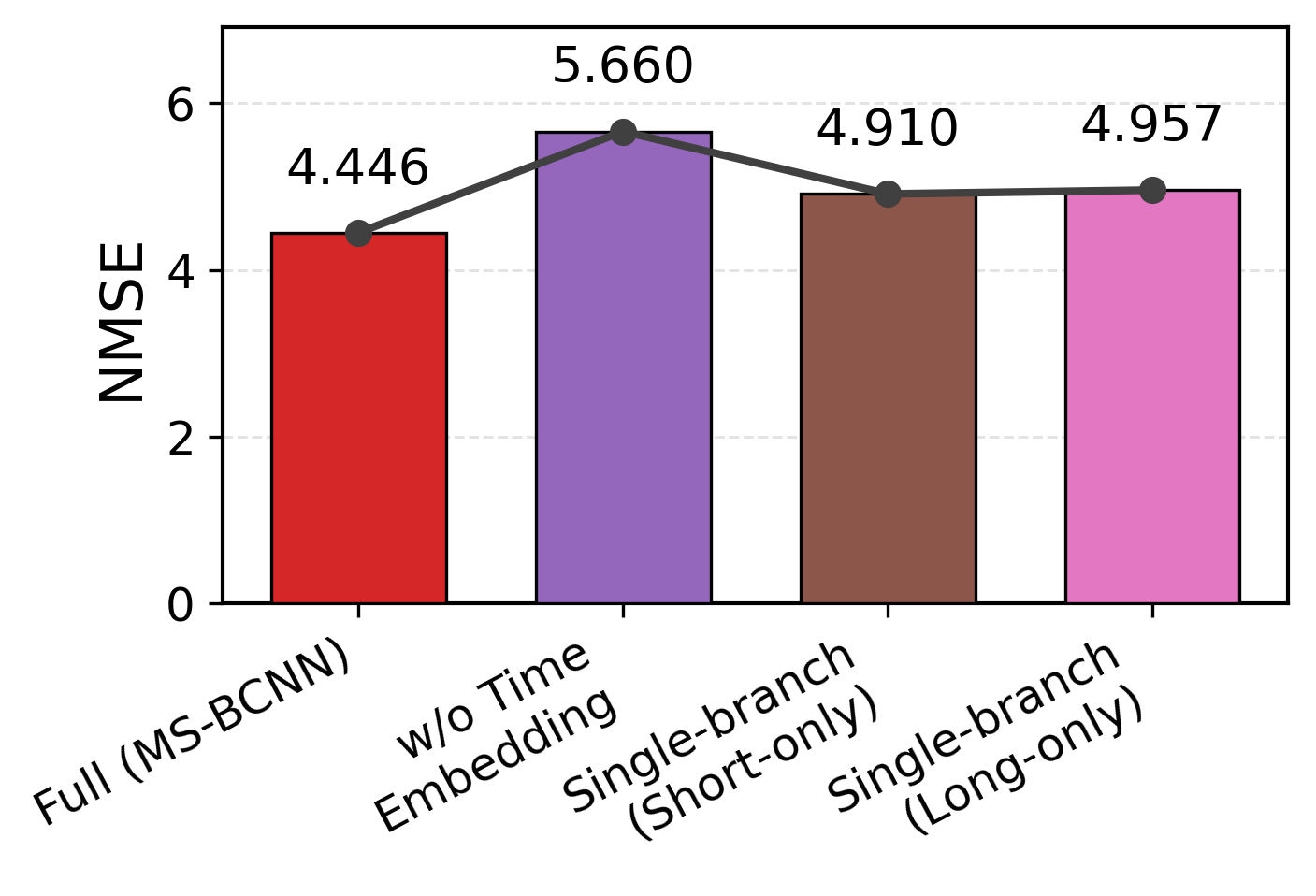}
    \\[0.2em]
    \small (b) MS-BCNN ablation
  \end{minipage}

  \caption{Ablation results of the proposed methods. Panel (a) reports IGD for PHMOEA, averaged over two synthetic benchmarks (H-DTLZ2 and H-DTLZ7), while panel (b) reports NMSE on the real-world sintering dataset for MS--BCNN. For both metrics, lower values indicate better performance.}
  \label{fig:ablation_igd_nmse}
\end{figure}

To assess the contribution of key components, we conduct ablation studies for both PHMOEA and MS--BCNN.
Fig.~\ref{fig:ablation_igd_nmse} summarizes the overall results, where panel (a) reports the mean IGD averaged over H-DTLZ2 and H-DTLZ7.

First, for PHMOEA, removing elitism leads to the most pronounced performance degradation, indicating that elitist preservation is crucial for stable convergence (Fig.~\ref{fig:ablation_igd_nmse}a).
In contrast, removing de-duplication causes only a mild drop, consistent with its role in improving efficiency by avoiding redundant evaluations; detailed statistics and IGD trajectories are provided in Appendix Table~\ref{tab:abl_phmoea_app} and Appendix Fig.~\ref{fig:app_igd_curves}.

Second, for MS--BCNN, removing the time embedding noticeably worsens the overall NMSE, suggesting that explicit temporal cues are critical for chronological forecasting (Fig.~\ref{fig:ablation_igd_nmse}b).
In addition, both the short-only and long-only variants underperform the dual-branch design, supporting the benefit of multi-scale modeling for real-world dynamic sintering forecasting; per-target details are deferred to Appendix Table~\ref{tab:ablation_detail}.

\section{Conclusion and Future Work}
\label{Section6}

This paper studies multi-output quality prediction on sintering data with multi-source asynchronous sampling, focusing on deployment-constrained modeling and automatic configuration.
We present a deployment-oriented configurable modeling and search framework that unifies alignment operators, scale choices, and network structures in a hierarchical conditional space and jointly optimizes the error--complexity trade-off to obtain a deployable Pareto set.
Concretely, we use a multi-scale bi-branch convolutional backbone to capture short-term fluctuations and long-term trends, and develop a multi-objective evolutionary search strategy to efficiently explore configurations under a limited budget while reducing redundant evaluations.
Results on synthetic benchmarks and a real-world sintering dataset demonstrate that our framework yields better trade-off solutions than strong baselines and alternative search strategies, enabling more flexible model selection in deployment.

While this work focuses on deployable modeling and trade-off solution discovery, real-world sintering production often exhibits pronounced regime changes and distribution drift, which can erode predictive performance over time.
An important future direction is to develop dynamic learning models for such nonstationarity, enabling controlled updates and stable maintenance under drift, and further characterizing drift severity to support more fine-grained response strategies.
Building on this, the Pareto set produced in this paper can serve as candidate initializations and structural priors for subsequent dynamic learning, reducing adaptation cost and improving robustness in long-term operation.
We believe that coupling offline configurable search with online drift-responsive maintenance will improve the reliability and scalability of multi-source asynchronous industrial time-series forecasting in real deployments.

\bibliography{refs}
\bibliographystyle{icml2026}

\clearpage            
\onecolumn            
\appendix
\raggedbottom   
\section*{Appendix}
\section{Definition of the PHMOEA Search Space}

\subsection{Hierarchical decision variables and activation rules}
\label{app:hier-decision}

To formalize conditional activation, we model the configuration space as a hierarchical conditional mixed space, where the activity of some variables is determined by higher-level discrete choices.

The $i$-th candidate configuration is represented as a hierarchical decision vector over levels $\mathcal{L}$, where level $\ell$ contains $n_\ell$ variables:
\begin{equation}
x_i=\big(x_{i,j}^{(\ell)}\big)_{\ell\in\mathcal{L},\, j=1,\ldots,n_\ell}.
\label{eq:hier-x-app}
\end{equation}

For each hierarchical variable, we define an activation indicator $a_{j}^{(\ell)}(x)\in\{0,1\}$: the variable is active if and only if its parent decision is selected.
This induces the active-variable set $\mathcal{A}(x)$, which characterizes feasibility constraints:
\begin{equation}
\mathcal{A}(x)=\left\{(\ell,j)\mid a_{j}^{(\ell)}(x)=1\right\}.
\label{eq:active-set-app}
\end{equation}

In the implementation, inactive variables do not participate in model construction and training,
and are stored using placeholders in the encoding to keep a unified dimensionality.

\subsection{Trainable-Parameter Counting for Model Complexity}
\label{app:param-count}

This section specifies the counting protocol of $P(x)$ for measuring model complexity.
We count only trainable parameters in the network specified by $x$.

Parameters counted in $P(x)$ include the following.

\par\noindent(1) weights and biases of convolutional and linear layers.

\par\noindent(2) affine parameters in normalization layers (e.g., $\gamma$ and $\beta$ when \texttt{affine} is enabled).

\par\noindent(3) trainable parameters introduced by fusion modules (e.g., linear projections in gating, attention, or multi-head attention).

\par\noindent(4) learnable branch weights under weighted-sum fusion.

\par\noindent(5) weights and biases of the prediction head (MLP) and the output layer.

We do not count parameter-free operators (e.g., activations, pooling, and dropout) or non-trainable buffers (e.g., BatchNorm running mean and running variance).

\subsection{Per-generation objective and crowding normalization in PHMOEA}
\label{app:norm-objectives}

To mitigate scale differences between objectives during evolutionary search, we apply per-generation normalization to objectives and crowding in PHMOEA.
This normalization is used only for search-related weighting and sampling guidance, and does not affect the raw objective values reported in results.

Let $\mathcal{S}^{(t)}$ denote the set of evaluated individuals in generation $t$, and let $\varepsilon>0$ be a numerical stabilizer.
In this work, the two objectives are $f_1(x)=\mathrm{MSE}(x)$ and $f_2(x)=P(x)$.

We compute the minimum and maximum of each objective over $\mathcal{S}^{(t)}$:
\begin{equation}
f_{m,\min}^{(t)}=\min_{x\in\mathcal{S}^{(t)}} f_m(x),\quad
f_{m,\max}^{(t)}=\max_{x\in\mathcal{S}^{(t)}} f_m(x),
\qquad m\in\{1,2\}.
\label{eq:fminmax-app}
\end{equation}

We then define the per-generation normalized objectives (denoted as $\hat{f}_m(x)$ in the main text):
\begin{equation}
\hat{f}_m(x)=
\frac{f_m(x)-f_{m,\min}^{(t)}}
{f_{m,\max}^{(t)}-f_{m,\min}^{(t)}+\varepsilon},
\qquad m\in\{1,2\}.
\label{eq:obj-norm-app}
\end{equation}

The crowding measure $c_i$ is computed as the standard NSGA-II crowding distance and normalized within the generation to obtain $\hat{c}_i$:
\begin{equation}
\hat{c}_i=
\frac{c_i-c_{\min}^{(t)}}{c_{\max}^{(t)}-c_{\min}^{(t)}+\varepsilon},
\qquad
c_{\min}^{(t)}=\min_{x_i\in\mathcal{S}^{(t)}}c_i,\ 
c_{\max}^{(t)}=\max_{x_i\in\mathcal{S}^{(t)}}c_i .
\label{eq:crowd-norm-app}
\end{equation}

This normalization is a monotone affine transformation within each generation and therefore does not change Pareto dominance within that generation.

\subsection{Configuration space specification}
\label{app:search-space}

To facilitate reproducibility and future extensions, we provide the complete list of decision variables and their conditional dependencies used in PHMOEA.
In our case study, we use a fixed-length encoding and represent a candidate configuration as $x=(x_1,\ldots,x_{24})$, where inactive conditional variables are kept as placeholders.
A conditional variable is active only when its parent discrete choice satisfies the condition in Table~\ref{tab:decision-space}; this can be indicated by $a_k(x)\in\{0,1\}$, and we mask variables with $a_k(x)=0$ during decoding to keep configurations executable.
Table~\ref{tab:decision-space} summarizes variable names, types, ranges and candidate sets.

We encode sample-level time variables using sine--cosine periodic functions to obtain a time-embedding vector; this is a standard component and is summarized in the main text, while we provide the exact definition here for reproducibility.
\begin{equation}
\boldsymbol{z}^{\mathrm{time}}
=
\mathrm{Concat}_{k=1}^{K_t}
\Big[
\sin\!\Big(2\pi \tfrac{\tau_k}{P_k}\Big),\ 
\cos\!\Big(2\pi \tfrac{\tau_k}{P_k}\Big)
\Big].
\label{eq:time-embed}
\end{equation}

Here $\tau_k$ and $P_k$ denote the $k$-th time variable and its natural period, respectively, $K_t$ is the number of time variables, and $\mathrm{Concat}(\cdot)$ denotes concatenation along the feature dimension.

For each source $j\in\{1,\ldots,S\}$, let the aligned sequence be $\tilde{\mathbf{X}}^{(j)}\in\mathbb{R}^{L_p\times C_j}$; we concatenate features from all sources along the channel dimension and append the time embedding broadcast along the time axis to obtain the final aligned input $\tilde{\mathbf{X}}$.
\begin{equation}
\tilde{\mathbf{X}}
=
\mathrm{Concat}\!\left(
\tilde{\mathbf{X}}^{(1)},\ldots,\tilde{\mathbf{X}}^{(S)},\
\mathbf{1}_{L_p}\,(\boldsymbol{z}^{\mathrm{time}})^{\top}
\right)
\in
\mathbb{R}^{L_p\times C_{\mathrm{in}}},
\label{eq:concat-input}
\end{equation}
Here $\mathbf{1}_{L_p}$ is an all-ones vector of length $L_p$ used to replicate the sample-level embedding along the time axis, and $C_{\mathrm{in}}=\sum_{j=1}^{S} C_j + 2K_t$ is the resulting input channel dimension.

We next specify the deterministic computation graph used to instantiate MS--BCNN from a configuration in Table~\ref{tab:decision-space}.

Let $\mathbf{H}^{(0)}\in\mathbb{R}^{L_p\times C_0}$ be the projected feature sequence, where $C_0=x_6$.

We use a fixed 3-layer bi-branch 1D convolutional stack with channel widths $C_1=x_7$, $C_2=x_8$, and $C_3=x_9$; the short and long branches share the same channel width at each layer and differ only in kernel sizes specified by $x_{10}$ and $x_{11}$.

All convolutions use stride $1$ and length-preserving padding (for odd kernel size $k$, we set $p=(k-1)/2$), so the temporal length remains $L_p$ throughout the backbone.

Within each branch and each layer, the operator order is fixed as $\mathrm{Conv1D}\rightarrow \mathrm{Norm}(x_5)\rightarrow \mathrm{Act}(x_{12})\rightarrow \mathrm{Dropout}(x_{13})$; no additional pooling or residual connections are used in the backbone.

After the 3rd layer, we obtain $\mathbf{H}^{\mathrm{short}},\mathbf{H}^{\mathrm{long}}\in\mathbb{R}^{L_p\times C_f}$ with $C_f=C_3=x_9$ and fuse them using $x_{22}$ to get $\mathbf{H}\in\mathbb{R}^{L_p\times C_f}$.
We directly flatten $\mathbf{H}$ into $\mathbf{f}\in\mathbb{R}^{L_p C_f}$ (without global average pooling) and apply a single linear prediction head
\begin{equation}
\hat{\mathbf{y}}=\mathbf{W}\mathbf{f}+\mathbf{b},\quad \mathbf{W}\in\mathbb{R}^{K\times (L_p C_f)},
\end{equation}
where $K$ is the number of prediction targets.

\begin{table}[H]
  \centering
  \small
  \caption{Decision variables in PHMOEA.}
  \label{tab:decision-space}
  \setlength{\tabcolsep}{4pt}
  \renewcommand{\arraystretch}{1.12}
  \begin{tabularx}{\textwidth}{@{}l >{\raggedright\arraybackslash}X l >{\raggedright\arraybackslash}X@{}}
    \toprule
    Variable & Name & Type & Range / Candidate Set \\
    \midrule
    $x_1$  & Resampling operator $\mathcal{R}$ & Discrete &
    \{\texttt{linear}, \texttt{decimate\_repeat}, \texttt{hybrid}, \texttt{pool}, \texttt{conv\_blurpool}, \texttt{fir\_lowpass}\} \\

    $x_2$  & Pooling type (active if $x_1=\texttt{pool}$) & Conditional Discrete &
    \{\texttt{avg}, \texttt{max}, \texttt{median}, \texttt{weighted}\} \\

    $x_3$  & Aligned length $L_p$ & Discrete &
    \{8, 12, 24, 36, 48\} \\

    $x_4$  & Batch size & Discrete &
    \{16, 32, 64, 128\} \\

    $x_5$  & Normalization layer & Discrete &
    \{\texttt{BatchNorm}, \texttt{LayerNorm}, \texttt{InstanceNorm}\} \\

    $x_6$  & Projection channels $C_0$ & Discrete &
    \{8, 16, 32, 64\} \\

    $x_7$  & Channels (conv1) & Discrete &
    \{8, 16, 32, 64\} \\

    $x_8$  & Channels (conv2) & Discrete &
    \{16, 32, 64, 128\} \\

    $x_9$  & Channels (conv3) & Discrete &
    \{32, 64, 128, 256\} \\

    $x_{10}$  & Short-branch kernel sizes (3 layers) & Discrete &
    \{(3,3,3), (3,5,7), (3,5,9), (5,7,11)\} \\

    $x_{11}$  & Long-branch kernel sizes (3 layers) & Discrete &
    \{(7,9,11), (9,11,13), (11,13,15)\} \\

    $x_{12}$  & Activation & Discrete &
    \{\texttt{ReLU}, \texttt{GELU}, \texttt{SiLU}, \texttt{Tanh}\} \\

    $x_{13}$ & Dropout rate & Continuous &
    $[0, 0.5]$ \\

    $x_{14}$ & Learning rate & Continuous &
    log-uniform in $[10^{-5}, 10^{-2}]$ \\

    $x_{15}$ & Weight decay & Continuous &
    log-uniform in $[10^{-6}, 10^{-2}]$ \\

    $x_{16}$ & Learning-rate scheduler & Discrete &
    \{\texttt{on}, \texttt{off}\} \\

    $x_{17}$ & Scheduler type (active if $x_{16}=\texttt{on}$) & Conditional Discrete &
    \{\texttt{plateau}, \texttt{warmup\_cosine}\} \\

    $x_{18}$ & Loss type & Discrete &
    \texttt{MSE}, \texttt{MAE}, \texttt{SmoothL1}, \texttt{MAPE}, \texttt{Huber}, \texttt{LogCosh}, \texttt{Quantile}, \texttt{SMAPE},\newline
    \texttt{Combined}, \texttt{AdaptiveCombined} \\

    $x_{19}$ & Combined-loss form (active if $x_{18}\in\{\texttt{Combined},\texttt{AdaptiveCombined}\}$) & Conditional Discrete &
    \{(\texttt{MSE},\texttt{MAE}), (\texttt{MSE},\texttt{Huber}), (\texttt{MAE},\texttt{Huber}), (\texttt{MAE},\texttt{MAPE}),\newline
    (\texttt{MSE},\texttt{SMAPE}), (\texttt{MAE},\texttt{Quantile}), (\texttt{Huber},\texttt{Quantile}), (\texttt{MAE},\texttt{multi\_quantile}),\newline
    (\texttt{MSE},\texttt{SmoothL1}), (\texttt{MAE},\texttt{LogCosh})\} \\

    $x_{20}$ & Loss weights (active if $x_{18}=\texttt{AdaptiveCombined}$) & Conditional Discrete &
    \{(0.9,0.1), (0.7,0.3), (0.5,0.5)\} \\

    $x_{21}$ & Learning rate for combined loss (active if $x_{18}=\texttt{AdaptiveCombined}$) & Conditional Discrete &
    log-spaced candidates \{0.001, 0.01, 0.1, 0.2, 0.5\} \\

    $x_{22}$  & Fusion operator $\mathrm{Fuse}_{\mathrm{conv}}$ & Discrete &
    \{\texttt{concat}, \texttt{add}, \texttt{weighting}, \texttt{gating}, \texttt{attention}, \texttt{cross\_mapping}\} \\

    $x_{23}$ & Weighting mode (active if $x_{22}=\texttt{weighting}$) & Conditional Discrete &
    \{\texttt{add}, \texttt{concat}\} \\

    $x_{24}$ & Cross-mapping mode (active if $x_{22}=\texttt{cross\_mapping}$) & Conditional Discrete &
    \{\texttt{add}, \texttt{concat}, \texttt{gated}\} \\
    \bottomrule
  \end{tabularx}
\end{table}

\subsection{Encoding and decoding details}
\label{app:encdec}

For the $d$-th continuous variable with range $[a_d,b_d]$, we discretize it into $K_d$ bins and use each bin midpoint as the representative value (linear scale):
\begin{equation}
v_{d,k} = a_d + \frac{2k - 1}{2K_d}\,(b_d - a_d),
\qquad k = 1,\dots,K_d .
\label{eq:linear-encoding-app}
\end{equation}

When a variable is positive and spans orders of magnitude (e.g., learning rate and weight decay), we use log-scale discretization by partitioning uniformly in log space and mapping back to the original space:
\begin{equation}
\begin{aligned}
v_{d,k}
&=
\exp\!\left(
  (1-\alpha_{d,k})\,\log(a_d)
  + \alpha_{d,k}\,\log(b_d)
\right),\\
\alpha_{d,k}
&=
\frac{2k - 1}{2K_d}.
\end{aligned}
\label{eq:log-encoding-app}
\end{equation}

For conditional variables, decoding activates them only when their parent discrete choices are selected (i.e., the corresponding activity indicator equals 1); otherwise they are masked and ignored during instantiation to ensure executability.

\subsection{Deduplication and retry-budget protocol}
\label{app:dedup}

Executable-configuration canonicalization: deduplication is performed after decoding and repair and before training evaluation; we retain only active variables and build a deterministic representation in a fixed order.
For a decoded candidate $x$, let $\mathcal{A}(x)$ denote its active-variable set, and denote the canonical form by $\mathrm{canon}(x)$.
We construct $\mathrm{canon}(x)$ as an ordered sequence of key--value pairs $\big((j,\mathrm{val}_j)\big)_{j\in\mathcal{A}(x)}$ sorted by dimension index: discrete variables use candidate IDs, while continuous variables use their bin indices to avoid spurious differences due to floating-point precision.

Hash key: we compute a hash key $\mathrm{key}(x)$ from the deterministic serialization $\mathrm{ser}(\mathrm{canon}(x))$ and check membership in a global key set $\mathcal{K}$.
Within a single run, $\mathcal{K}$ is maintained across generations and stores keys of configurations admitted to evaluation; once a candidate passes the check and is admitted, we insert its key into $\mathcal{K}$.

Retry budget: for each offspring slot, we attempt at most $n_{\text{trial}}$ trials to obtain a feasible and non-duplicated candidate; if all trials fail, we skip the slot and proceed to the next one to avoid stalling.
Because of the skip rule, the number of individuals admitted to evaluation in a generation can be smaller than the nominal $N$, consistent with the $N_t$ notation in the main-text complexity analysis.

\subsection{Adaptive refinement protocol for continuous variables}
\label{app:refine}

This appendix provides implementation details for adaptive refinement of continuous variables, including interval breakpoint construction, interval-mass statistics, and the update and trigger rules of the cross-generation state $\mathcal{S}$.

Breakpoints: for the $d$-th continuous variable at level $\ell$ with current range $[a_d^{(\ell)}, b_d^{(\ell)}]$, we uniformly split it into $K_d^{(\ell)}$ sub-intervals and define the breakpoints as:
\begin{equation}
\beta^{(\ell)}_{d,k}
=
a_d^{(\ell)} + \frac{k}{K_d^{(\ell)}}\big(b_d^{(\ell)}-a_d^{(\ell)}\big),
\qquad k=0,1,\ldots,K_d^{(\ell)}.
\label{eq:refine-boundary}
\end{equation}

Intervals and bin assignment: the $k$-th interval is $I^{(\ell)}_{d,k}=[\beta^{(\ell)}_{d,k-1},\,\beta^{(\ell)}_{d,k})$ (with the last interval right-closed to include the boundary), and we assign a decoded value (or its bin index) to an interval accordingly.

Interval mass: let $P_t^\star$ be the current generation's non-dominated set; we define the interval mass as
\begin{equation}
f^{(\ell)}_{d,k}(t)
=
\frac{1}{|P_t^\star|}
\sum_{x\in P_t^\star}
\mathbb{I}\!\left[x_{d}^{(\ell)} \in I^{(\ell)}_{d,k}\right],
\label{eq:interval-mass}
\end{equation}

where $\mathbb{I}[\cdot]$ is the indicator function and $x_{d}^{(\ell)}$ denotes the decoded value (or equivalently, the bin assignment) of individual $x$ on this continuous variable.

Cross-generation state and trigger: for each $(\ell,d,k)$ we maintain a consecutive counter $c^{(\ell)}_{d,k}(t)\in\{0,1,\ldots\}$ and trigger refinement with threshold $\delta_h$ and persistence window $H$:
\begin{equation}
c^{(\ell)}_{d,k}(t)
=
\begin{cases}
c^{(\ell)}_{d,k}(t-1)+1, & f^{(\ell)}_{d,k}(t)>\delta_h,\\
0, & \text{otherwise},
\end{cases}
\label{eq:persistence-counter}
\end{equation}
When $c^{(\ell)}_{d,k}(t)\ge H$, we refine the interval $I^{(\ell)}_{d,k}$ (e.g., by further splitting it into finer sub-intervals and updating the corresponding partition for dimension $(\ell,d)$).

Representative values: after refinement, representative values used for encoding and decoding follow the same unified discretization; the linear/log mappings are given in Appendix~\ref{app:encdec}.

\subsection{Early-Stopping Mechanism}
\label{app:early-stop}

To reduce redundant expensive function evaluations after the Pareto front has largely converged, we check an early-stopping rule before generating/evaluating offspring in each generation; if progress has been stagnant over the last $W$ generations, we terminate at $T_{\mathrm{stop}}<T_{\max}$.

The rule depends only on generation-wise statistics of the bi-objective minimization $F(x)=(f_1(x),f_2(x))$ and can be applied uniformly to synthetic benchmarks, surrogate benchmarks, and the real-world task.

Let $P_t$ denote the population at generation $t$, and let $P_t^\star \subseteq P_t$ denote its current non-dominated set (the first front).

We define the mean objective value on $P_t^\star$ (for $i\in\{1,2\}$) as:
\begin{equation}
\overline{f}_i(t)=\frac{1}{|P_t^\star|}\sum_{x\in P_t^\star} f_i(x).
\label{eq:earlystop-front-mean}
\end{equation}

To avoid scale sensitivity, we use a relative-improvement convention with a window size $W$:
\begin{equation}
\Delta \overline{f}_i(t)
=
\frac{\max\!\left\{0,\ \overline{f}_i(t-W)-\overline{f}_i(t)\right\}}
{\max\!\left\{\left|\overline{f}_i(t-W)\right|,\ \epsilon_0\right\}},
\quad i\in\{1,2\}.
\label{eq:earlystop-delta-fi}
\end{equation}

Front coverage is measured by hypervolume $HV(P_t^\star)$ (defined in Appendix C.4) with the relative gain:
\begin{equation}
\Delta HV(t)
=
\frac{\max\!\left\{0,\ HV(P_t^\star)-HV(P_{t-W}^\star)\right\}}
{\max\!\left\{\left|HV(P_{t-W}^\star)\right|,\ \epsilon_0\right\}}.
\label{eq:earlystop-delta-hv}
\end{equation}
We set the HV reference point to $r=(r_1,r_2)$ with $r_i=1.1\cdot \max_{x\in P_0} f_i(x)$ for $i\in\{1,2\}$, and keep $r$ fixed for all subsequent generations.

Before generating/evaluating offspring at each generation, we compute the three stagnation signals in Eqs.~(\ref{eq:earlystop-delta-fi})--(\ref{eq:earlystop-delta-hv}); early stopping is triggered only when all of them fall below their thresholds.

In all experiments, we use $W=8$, $\epsilon_0=10^{-12}$, $\epsilon_1=10^{-3}$, $\epsilon_2=10^{-3}$, and $\epsilon_{HV}=10^{-4}$, where $\epsilon_1$ and $\epsilon_2$ are thresholds for $\Delta \overline{f}_1(t)$ and $\Delta \overline{f}_2(t)$, respectively.

\subsection{Complexity details}
\label{app:complexity}

This subsection supplements the main-text complexity statement by clarifying the counting protocol and outlining the derivation, and by bounding non-evaluation overhead to show it does not change the dominant term.

The population size is $N$ and the algorithm runs for at most $T_{\max}$ generations; one function evaluation (FE) corresponds to one full train--validate run of a candidate configuration that returns its objective vector.

Let $N_t$ denote the number of candidates admitted to evaluation at generation $t$. Because feasibility repair and deduplication are applied before evaluation and an offspring slot can be skipped when the retry budget is exhausted, we have $0\le N_t\le N$.

If the algorithm terminates at $T_{\mathrm{stop}}\le T_{\max}$ generations, the total number of evaluations is
\begin{equation}
\mathrm{FEs}=N+\sum_{t=1}^{T_{\mathrm{stop}}}N_t,
\label{eq:fe-sum}
\end{equation}
where the leading term $N$ accounts for evaluating the initial population $P_0$.

Let $C$ be the average cost of one train--validate run. The total evaluation cost is $O(\mathrm{FEs}\cdot C)$; in the worst case where $N_t=N$ and $T_{\mathrm{stop}}=T_{\max}$, this reduces to $O(T_{\max}NC)$.

For non-evaluation overhead, per-generation operations such as non-dominated sorting, archive updates, sampling, and repair/deduplication scale at most linearly or near-linearly with $N$ (e.g., $O(N\log N)$ or $O(NJ)$). Compared with the training cost $C$, these terms are auxiliary and do not change the dominant complexity.

\section{Experimental Details}

\subsection{Real-world Sintering Dataset}
\label{app:sinter-dataset}

This section summarizes the input--target specification and the preprocessing/splitting protocol of the real-world sintering dataset for reproducibility and for preventing temporal leakage.
Table~\ref{tab:full-features} lists all input variables and output targets, grouped by category.

\begin{table}[H]
  \centering
  \small
  \setlength{\tabcolsep}{4pt}
  \renewcommand{\arraystretch}{1.05}
  \caption{Input variables and output targets in the sintering dataset (grouped by category).}
  \label{tab:full-features}
  \begin{tabularx}{\linewidth}{p{4.2cm} >{\raggedright\arraybackslash}X}
    \toprule
    Category & Variables / Targets \\
    \midrule
    Raw-material composition and blending &
    Mixed-material moisture, temperature, and granularity;
    fuel granularity; coarse fraction ratio (particle size $>5$\,mm);
    ore-type proportions in the blend (six ore types);
    second-mixing additions (e.g., quicklime, return fines, coke breeze);
    planned blending ratios; planned basicity;
    raw-material price indicators (five series). \\
    \midrule
    Process and thermal &
    Temperatures (ignition, exhaust gas, BTP);
    pressures (furnace negative pressure, main-line negative pressure, gas pressure, air pressure);
    flows (coal-gas flow, air flow);
    ore-bin scale flow rates (six bins). \\
    \midrule
    Equipment state and environment &
    Material bed thickness; plate-belt speed; sintering machine speed;
    finished-product weighing; production output. \\
    \midrule
    Output targets &
    TFe; FeO; SiO$_2$; CaO; Basicity. \\
    \bottomrule
  \end{tabularx}
\end{table}

We summarize the preprocessing and split protocol below, following filtering, missing-value repair, cycle alignment, normalization, and splitting.

(1) Sample filtering: we remove samples from non-stationary periods such as startup and shutdown to reduce the impact of transient regimes on modeling and evaluation.

(2) Missing-value repair: we apply time-ordered linear interpolation only to short missing gaps; longer gaps are not interpolated and are handled by the missing-data policy.

Time variables and time embedding: for each cycle-level sample, we extract calendar components from its associated cycle timestamp and build the time embedding by Eq.~(\ref{eq:time-embed}).

We fix the cycle timestamp as the start time of the corresponding production cycle and use four components (year, month, day-of-month, and hour), hence $K_t=4$.

We use zero-based indices and define $\{(\tau_k,P_k)\}_{k=1}^{K_t}$ deterministically as:
$\tau_1=\mathrm{year}-\mathrm{year}_{\min}$ and $P_1=\mathrm{year}_{\max}-\mathrm{year}_{\min}+1$;
$\tau_2=\mathrm{month}-1$ and $P_2=12$;
$\tau_3=\mathrm{day}-1$ (day-of-month) and $P_3=31$;
$\tau_4=\mathrm{hour}$ and $P_4=24$.

With zero-based indices, $\tau_k\in[0,P_k-1]$ holds by construction and no additional modulo operation is applied; we directly use Eq.~(\ref{eq:time-embed}) without further normalization of $\tau_k$.

(4) Normalization and leakage prevention: we apply per-feature Min--Max scaling; statistics are estimated on the training split only and then fixed for transforming validation and test splits to prevent temporal leakage.

(5) Data splits: we use a chronological train/validation/test split (0.70/0.15/0.15) to mimic forward-in-time deployment, and additionally run a control evaluation on a randomly shuffled split to approximate an independent and identically distributed (i.i.d.) setting; both settings use the same split ratio and evaluation protocol.

\subsection{Hierarchical Synthetic Benchmarks}
\label{app:benchmarks-hdtlz}

We construct two bi-objective ($M=2$) hierarchical synthetic benchmarks based on DTLZ2 and DTLZ7. The hierarchical structure is realized by a projection operator $z=\Pi(x)$ that decodes a mixed decision $x$ and extracts currently active continuous variables to form $z=\Pi(x)\in[0,1]^n$. Inactive continuous variables are set to a problem-specific neutral value consistent with the Pareto set so that the reference Pareto front remains reachable. In particular, we set inactive variables to $0.5$ for H-DTLZ2 and to $0$ for H-DTLZ7.

We introduce a hierarchical coupling regularizer on tail variables to emphasize parent--child dependencies. Let $a_j(x)\in\{0,1\}$ denote the activity indicator of the $j$-th continuous variable, and define the active tail-index set as $\mathcal{J}_c(x)=\{j\in\{3,\ldots,n\}\mid a_j(x)=1\}$. The coupling term is defined as
\begin{equation}
c(z,x)=
\begin{cases}
0, & \lvert \mathcal{J}_c(x)\rvert = 0,\\[0.3em]
\frac{1}{\lvert \mathcal{J}_c(x)\rvert}\sum\limits_{j\in\mathcal{J}_c(x)}\big(z_j-z_{p(j)}\big)^2, & \lvert \mathcal{J}_c(x)\rvert > 0,
\end{cases}
\qquad
g_{\mathrm{cpl}}(z,x)=\gamma\,c(z,x),
\end{equation}
where $\gamma\ge 0$ is a coupling coefficient.

To instantiate the parent index $p(j)$ for $j\in\{3,\ldots,n\}$, we consider two topologies and define two parent-index functions:
\begin{equation}
p_{\mathrm{chain}}(j)=j-1,\qquad
p_{\mathrm{tree}}(j)=\left\lfloor\frac{j-3}{2}\right\rfloor+2,
\qquad j\in\{3,\ldots,n\}.
\end{equation}
We use $p(j)=p_{\mathrm{chain}}(j)$ for the chain topology and $p(j)=p_{\mathrm{tree}}(j)$ for the binary-tree topology.

\paragraph{H-DTLZ2 (normalized and coupled, $M=2$).}
For H-DTLZ2, we use the standard DTLZ2 angular mapping and define
\begin{equation}
g_2(z,x)=\frac{1}{n-1}\sum_{j=2}^{n}(z_j-0.5)^2 + g_{\mathrm{cpl}}(z,x),
\end{equation}
\begin{equation}
f_1(z,x)=(1+g_2(z,x))\cos\!\left(\frac{\pi}{2}z_1\right),\qquad
f_2(z,x)=(1+g_2(z,x))\sin\!\left(\frac{\pi}{2}z_1\right),
\end{equation}
and
\begin{equation}
F_{\mathrm{H\text{-}DTLZ2}}(x)=\big(f_1(\Pi(x),x),f_2(\Pi(x),x)\big).
\end{equation}
The Pareto front satisfies $g_2(z,x)=0$ and remains the quarter unit circle in the first quadrant:
\begin{equation}
\mathcal{P}^{\star}_{\mathrm{H\text{-}DTLZ2}}
=\left\{\big(\cos(\tfrac{\pi}{2}u),\,\sin(\tfrac{\pi}{2}u)\big)\ \big|\ u\in[0,1]\right\}.
\end{equation}

\paragraph{H-DTLZ7 (normalized and coupled, $M=2$).}
For H-DTLZ7, we define
\begin{equation}
f_1(z)=z_1,\qquad
g_7(z,x)=1+\frac{9}{n-1}\sum_{j=2}^{n} z_j + g_{\mathrm{cpl}}(z,x),
\end{equation}
\begin{equation}
h(z,x)=2-\frac{f_1(z)}{g_7(z,x)}\left(1+\sin\big(3\pi f_1(z)\big)\right),\qquad
f_2(z,x)=\frac{1}{2}\,g_7(z,x)\,h(z,x),
\end{equation}
and
\begin{equation}
F_{\mathrm{H\text{-}DTLZ7}}(x)=\big(f_1(\Pi(x)),f_2(\Pi(x),x)\big).
\end{equation}
For $M=2$, the Pareto front under $g_7(z,x)=1$ can be defined via non-dominated filtering of the parameterized candidate set:
\begin{equation}
\mathcal{P}^{\star}_{\mathrm{H\text{-}DTLZ7}}
=
\mathrm{ND}\!\left(
\left\{
\left(u,\,
\frac{2-u\big(1+\sin(3\pi u)\big)}{2}
\right)
\ \middle|\ u\in[0,1]
\right\}
\right),
\end{equation}
where $\mathrm{ND}(\cdot)$ denotes the non-dominated filtering operator under minimization.

\subsection{Decision Variables and Operator Definitions}
\label{app:resampling-ops}

This section is aligned with Table~\ref{tab:decision-space}: following $x_1$ to $x_{24}$, we clarify the meaning of each decision variable and provide operator and loss definitions when needed (candidate sets and ranges are given in Table~\ref{tab:decision-space}) to ensure reproducibility.

Let a source sequence before alignment be $\mathbf{X}\in\mathbb{R}^{T\times F}$ and the aligned output be $\tilde{\mathbf{X}}\in\mathbb{R}^{L_p\times F}$.
In our search space, $L_p\ge 2$, so the continuous-time coordinate below is well-defined.
We define the continuous-time coordinate $u_t=\frac{t}{L_p-1}(T-1)$ for $t=0,\ldots,L_p-1$.
Let $[\![u]\!]$ denote rounding to the nearest integer and clipping to $[0,T-1]$ (implemented as $\min(T-1,\max(0,\lfloor u+0.5\rfloor))$).
To be consistent with Eq.~(\ref{eq:mse}), all losses below use normalized mean reduction over samples and output dimensions: for any set of $N_D$ pairs $\{(\mathbf{y}_n,\hat{\mathbf{y}}_n)\}_{n=1}^{N_D}$ with $K$ outputs ($\mathbf{y}_n,\hat{\mathbf{y}}_n\in\mathbb{R}^{K}$), we compute $\frac{1}{N_DK}\sum_{n=1}^{N_D}\sum_{k=1}^{K}\ell(y_{n,k},\hat y_{n,k})$ (for division-based losses such as MAPE/SMAPE, we add $\epsilon>0$ to denominators to avoid zero division).

\paragraph{$x_1$: Resampling operator $\mathcal{R}$.}
$x_1$ selects the resampling operator that maps $\mathbf{X}$ to length $L_p$.

\begin{enumerate}
  \renewcommand{\labelenumi}{(\arabic{enumi})}

  \item \texttt{linear} (linear interpolation).
  Define $i_t=\lfloor u_t\rfloor$ and $\lambda_t=u_t-i_t$.
  \begin{equation}
    \tilde{\mathbf{X}}[t,:]=
    \begin{cases}
      (1-\lambda_t)\mathbf{X}[i_t,:]+\lambda_t\mathbf{X}[i_t+1,:], & i_t<T-1,\\
      \mathbf{X}[T-1,:], & i_t=T-1.
    \end{cases}
  \end{equation}

  \item \texttt{decimate\_repeat} (uniform decimation for downsampling, repetition for upsampling).
  For downsampling, use $j_t=[\![u_t]\!]$.
  \begin{equation}
    j_t=[\![u_t]\!],\qquad \tilde{\mathbf{X}}[t,:]=\mathbf{X}[j_t,:].
  \end{equation}
  For upsampling, repeat samples to reach length $L_p$.
  The repetition counts are determined by $(q,s)$ with $q=\lfloor L_p/T\rfloor$ and $s=L_p-qT$.
  \begin{equation}
    n_i=
    \begin{cases}
      q+1,& i<s,\\
      q,& i\ge s,
    \end{cases}
    \qquad i=0,\ldots,T-1.
  \end{equation}
  Let $L_{-1}=0$ and $L_i=\sum_{k=0}^{i}n_k$. When $L_{i-1}\le t<L_i$, set $\tilde{\mathbf{X}}[t,:]=\mathbf{X}[i,:]$.

  \item \texttt{hybrid} (decimation + smoothing for downsampling, repetition/linear for upsampling).
  For downsampling, first uniformly decimate with $j_t=[\![u_t]\!]$ to obtain $\mathbf{Z}$, then apply three-point moving-average smoothing.
  \begin{equation}
    \mathbf{Z}[t,:]=\mathbf{X}[j_t,:].
  \end{equation}
  \begin{equation}
    \tilde{\mathbf{X}}[t,:]=
    \begin{cases}
      \mathbf{Z}[0,:], & t=0,\\
      \frac{1}{3}\big(\mathbf{Z}[t-1,:]+\mathbf{Z}[t,:]+\mathbf{Z}[t+1,:]\big), & 1\le t\le L_p-2,\\
      \mathbf{Z}[L_p-1,:], & t=L_p-1.
    \end{cases}
  \end{equation}
  For upsampling, it degenerates to repetition when $T=1$, otherwise it uses \texttt{linear}.

  \item \texttt{pool} (interval pooling for downsampling, repetition for upsampling).
  When $T>L_p$, we partition the time axis into $L_p$ adjacent intervals and aggregate within each interval, where the aggregation operator is selected by the conditional variable $x_2$.
  The boundaries are $e_t=\lfloor \frac{tT}{L_p}\rfloor$ for $t=0,\ldots,L_p$, and the index set is $\mathcal{I}_t=\{k\mid e_t\le k<e_{t+1}\}$.
  When $T\le L_p$, upsampling follows the same repetition-padding rule as \texttt{decimate\_repeat}.

  \item \texttt{conv\_blurpool} (blur convolution + decimation for downsampling, interpolation + blur for upsampling).
  This operator applies 1D convolution along the time axis independently for each feature dimension, using reflective padding at the boundaries.
  For downsampling, blur with $h=\frac{1}{16}[1,4,6,4,1]$ and then uniformly decimate with $j_t=[\![u_t]\!]$.
  \begin{equation}
    h=\frac{1}{16}[1,4,6,4,1],\qquad \mathbf{Z}=\mathbf{X}*h,\qquad \tilde{\mathbf{X}}[t,:]=\mathbf{Z}[j_t,:].
  \end{equation}
  For upsampling, first obtain $\hat{\mathbf{X}}$ by \texttt{linear}, then blur with the same kernel $h$.
  \begin{equation}
    \hat{\mathbf{X}}=\texttt{linear}(\mathbf{X}),\qquad \tilde{\mathbf{X}}=\hat{\mathbf{X}}*h.
  \end{equation}

  \item \texttt{fir\_lowpass} (windowed-sinc FIR low-pass + decimation for downsampling, interpolation + mild FIR smoothing for upsampling).
  This operator performs filtering and decimation along the time axis independently for each feature dimension.
  For downsampling, we apply a windowed-sinc FIR low-pass kernel $h_{\mathrm{fir}}$ (Hamming window) for anti-aliasing and then uniformly decimate with $j_t=[\![u_t]\!]$.
  Let $d=T/L_p$ and $f_c\approx 0.5/d$.
  \begin{equation}
    \mathbf{Z}=\mathbf{X}*h_{\mathrm{fir}},\qquad \tilde{\mathbf{X}}[t,:]=\mathbf{Z}[j_t,:].
  \end{equation}
  For upsampling, we first obtain $\hat{\mathbf{X}}$ by \texttt{linear}, then apply a small mean FIR kernel $\bar{h}=\frac{1}{K}[1,\ldots,1]$ for mild smoothing (a small odd $K\le 5$ in our implementation).
  \begin{equation}
    \hat{\mathbf{X}}=\texttt{linear}(\mathbf{X}),\qquad \tilde{\mathbf{X}}=\hat{\mathbf{X}}*\bar{h}.
  \end{equation}

\end{enumerate}

\paragraph{$x_2$: Pooling type (active if $x_1=\texttt{pool}$).}
Active only when $x_1=\texttt{pool}$, using the interval partition $\mathcal{I}_t$ defined under $x_1$. Each interval is aggregated into one output point, computed independently per feature.

\begin{enumerate}
  \renewcommand{\labelenumi}{(\arabic{enumi})}

  \item \texttt{avg}:
  interval average pooling.
  \begin{equation}
    \tilde{\mathbf{X}}[t,:]=\frac{1}{|\mathcal{I}_t|}\sum_{k\in\mathcal{I}_t}\mathbf{X}[k,:].
  \end{equation}

  \item \texttt{max}:
  interval max pooling (element-wise over features).
  \begin{equation}
    \tilde{\mathbf{X}}[t,:]=\max_{k\in\mathcal{I}_t}\mathbf{X}[k,:].
  \end{equation}

  \item \texttt{median}:
  interval median pooling (element-wise over features).
  \begin{equation}
    \tilde{\mathbf{X}}[t,:]=\operatorname{median}\big(\{\mathbf{X}[k,:]\}_{k\in\mathcal{I}_t}\big).
  \end{equation}

  \item \texttt{weighted}:
  triangular-weighted average pooling with normalized weights within each interval, which degenerates to copying when $|\mathcal{I}_t|=1$.
  \begin{equation}
    \alpha_{t,k}=
    \begin{cases}
      1, & |\mathcal{I}_t|=1,\\[0.2em]
      1-\left|\dfrac{2(k-e_t)}{|\mathcal{I}_t|-1}-1\right|, & |\mathcal{I}_t|\ge 2,
    \end{cases}
    \qquad
    w_{t,k}=\dfrac{\alpha_{t,k}}{\sum_{j\in\mathcal{I}_t}\alpha_{t,j}}.
  \end{equation}
  \begin{equation}
    \tilde{\mathbf{X}}[t,:]=\sum_{k\in\mathcal{I}_t} w_{t,k}\,\mathbf{X}[k,:].
  \end{equation}

\end{enumerate}

\paragraph{$x_5$: Normalization layer.}
$x_5$ selects the normalization layer type using standard definitions with learnable affine parameters $\gamma,\beta$.
In the equations below, $\epsilon>0$ is a numerical-stability constant, and $\mu$ and $\sigma^2$ denote the mean and variance computed over the corresponding normalization axes.

\begin{enumerate}
  \renewcommand{\labelenumi}{(\arabic{enumi})}
  \item \texttt{BatchNorm}:
  \begin{equation}
    \mathrm{BN}(\mathbf{a})=\gamma\frac{\mathbf{a}-\mu_{\mathcal{B}}}{\sqrt{\sigma^2_{\mathcal{B}}+\epsilon}}+\beta.
  \end{equation}
  \item \texttt{LayerNorm}:
  \begin{equation}
    \mathrm{LN}(\mathbf{a})=\gamma\frac{\mathbf{a}-\mu_{\mathrm{feat}}}{\sqrt{\sigma^2_{\mathrm{feat}}+\epsilon}}+\beta.
  \end{equation}
  \item \texttt{InstanceNorm}:
  \begin{equation}
    \mathrm{IN}(\mathbf{a})=\gamma\frac{\mathbf{a}-\mu_{\mathrm{inst}}}{\sqrt{\sigma^2_{\mathrm{inst}}+\epsilon}}+\beta.
  \end{equation}
\end{enumerate}

\paragraph{$x_{12}$: Activation.}
$x_{12}$ selects a pointwise activation function. To avoid implementation ambiguity, we provide the standard mathematical definitions below, where $\sigma(\cdot)$ is the logistic sigmoid and $\Phi(\cdot)$ is the standard normal CDF.

\begin{enumerate}
  \renewcommand{\labelenumi}{(\arabic{enumi})}

  \item \texttt{ReLU}:
  \begin{equation}
    \mathrm{ReLU}(a)=\max(0,a).
  \end{equation}

  \item \texttt{GELU} (exact):
  \begin{equation}
    \mathrm{GELU}(a)=a\,\Phi(a),
    \qquad
    \Phi(a)=\int_{-\infty}^{a}\frac{1}{\sqrt{2\pi}}\exp\!\left(-\frac{t^2}{2}\right)\,dt .
  \end{equation}

  \item \texttt{SiLU}:
  \begin{equation}
    \mathrm{SiLU}(a)=a\,\sigma(a),
    \qquad
    \sigma(a)=\frac{1}{1+\exp(-a)} .
  \end{equation}

  \item \texttt{Tanh}:
  \begin{equation}
    \tanh(a)=\frac{\exp(a)-\exp(-a)}{\exp(a)+\exp(-a)} .
  \end{equation}
\end{enumerate}

\paragraph{$x_{16}$: Learning-rate scheduler.}
$x_{16}$ controls whether to enable learning-rate scheduling, and the specific scheduler is selected by the conditional variable $x_{17}$.

\paragraph{$x_{17}$: Scheduler type (active if $x_{16}=\texttt{on}$).}
When scheduling is enabled, $x_{17}$ selects the scheduler type.

\begin{enumerate}
  \renewcommand{\labelenumi}{(\arabic{enumi})}
  \item \texttt{plateau} (reduce-on-plateau).
  When the monitored metric does not improve for several epochs, the learning rate is multiplied by a fixed decay factor (standard implementation).

  \item \texttt{warmup\_cosine}.
  Let the total steps be $S$, the warmup steps be $S_w$, and the step index be $s=0,\ldots,S-1$. The learning rate is
  \begin{equation}
    \eta_s=
    \begin{cases}
      \eta_{\max}\dfrac{s+1}{S_w}, & 0\le s<S_w,\\[0.4em]
      \eta_{\min}+\dfrac{1}{2}(\eta_{\max}-\eta_{\min})\left(1+\cos\!\left(\pi\,\dfrac{s-S_w}{S-S_w}\right)\right), & S_w\le s\le S-1.
    \end{cases}
  \end{equation}
\end{enumerate}

\paragraph{$x_{18}$: Loss type.}
$x_{18}$ selects the training loss type. Below we use elementwise residual $r_{n,k}=y_{n,k}-\hat y_{n,k}$.

\begin{enumerate}
  \renewcommand{\labelenumi}{(\arabic{enumi})}

  \item \texttt{MSE}:
  \begin{equation}
    \mathcal{L}_{\mathrm{MSE}}=\frac{1}{N_DK}\sum_{n=1}^{N_D}\sum_{k=1}^{K} r_{n,k}^2.
  \end{equation}

  \item \texttt{MAE}:
  \begin{equation}
    \mathcal{L}_{\mathrm{MAE}}=\frac{1}{N_DK}\sum_{n=1}^{N_D}\sum_{k=1}^{K} |r_{n,k}|.
  \end{equation}

  \item \texttt{SmoothL1} (with a fixed transition $\beta>0$ in implementation):
  \begin{equation}
    \ell_{\mathrm{SmoothL1}}(r)=
    \begin{cases}
      \frac{1}{2\beta}r^2, & |r|<\beta,\\
      |r|-\frac{\beta}{2}, & \text{otherwise},
    \end{cases}
    \quad
    \mathcal{L}_{\mathrm{SmoothL1}}=\frac{1}{N_DK}\sum_{n,k}\ell_{\mathrm{SmoothL1}}(r_{n,k}).
  \end{equation}

  \item \texttt{MAPE} (with stabilizer $\epsilon>0$):
  \begin{equation}
    \mathcal{L}_{\mathrm{MAPE}}=\frac{100}{N_DK}\sum_{n,k}\frac{|r_{n,k}|}{|y_{n,k}|+\epsilon}.
  \end{equation}

  \item \texttt{Huber} (with a fixed threshold $\delta>0$ in implementation):
  \begin{equation}
    \ell_{\mathrm{Huber}}(r)=
    \begin{cases}
      \frac{1}{2}r^2, & |r|\le \delta,\\
      \delta\left(|r|-\frac{\delta}{2}\right), & |r|>\delta,
    \end{cases}
    \quad
    \mathcal{L}_{\mathrm{Huber}}=\frac{1}{N_DK}\sum_{n,k}\ell_{\mathrm{Huber}}(r_{n,k}).
  \end{equation}

  \item \texttt{LogCosh}:
  \begin{equation}
    \mathcal{L}_{\mathrm{LogCosh}}=\frac{1}{N_DK}\sum_{n,k}\log\!\cosh(r_{n,k}).
  \end{equation}

  \item \texttt{Quantile} (pinball loss with a fixed $\tau\in(0,1)$ in implementation):
  \begin{equation}
    \rho_{\tau}(r)=\max(\tau r,(\tau-1)r),
    \quad
    \mathcal{L}_{\mathrm{Quantile}}=\frac{1}{N_DK}\sum_{n,k}\rho_{\tau}(r_{n,k}).
  \end{equation}

  \item \texttt{SMAPE} (with stabilizer $\epsilon>0$):
  \begin{equation}
    \mathcal{L}_{\mathrm{SMAPE}}=\frac{100}{N_DK}\sum_{n,k}\frac{2|r_{n,k}|}{|y_{n,k}|+|\hat y_{n,k}|+\epsilon}.
  \end{equation}

  \item \texttt{Combined}.
  This option uses a two-term combined loss, where the loss pair is selected by the conditional variable $x_{19}$.
  \begin{equation}
    \mathcal{L}_{\mathrm{Combined}}=\alpha\,\mathcal{L}_a+(1-\alpha)\,\mathcal{L}_b,
  \end{equation}
  where $(\mathcal{L}_a,\mathcal{L}_b)$ is selected by $x_{19}$ and $\alpha$ is fixed to $0.5$ for a symmetric combination.

  \item \texttt{AdaptiveCombined}.
  This option also uses the loss pair selected by $x_{19}$, with weight initialization given by $x_{20}$ and weight-update learning rate given by $x_{21}$.
  \begin{equation}
    \mathcal{L}_{\mathrm{Adaptive}}=w_1\,\mathcal{L}_a+w_2\,\mathcal{L}_b,\qquad w_1,w_2\ge 0,\qquad w_1+w_2=1.
  \end{equation}

\end{enumerate}

\paragraph{$x_{22}$: Fusion operator $\mathrm{Fuse}_{\mathrm{conv}}$.}
$x_{22}$ selects the fusion operator for short/long branch features. Let $\mathbf{H}_s,\mathbf{H}_\ell\in\mathbb{R}^{L_p\times C}$ be the two branch outputs.

\begin{enumerate}
  \renewcommand{\labelenumi}{(\arabic{enumi})}

  \item \texttt{concat}:
  \begin{equation}
    \mathrm{Fuse}(\mathbf{H}_s,\mathbf{H}_\ell)=\left[\mathbf{H}_s\,\|\,\mathbf{H}_\ell\right]\in\mathbb{R}^{L_p\times 2C}.
  \end{equation}

  \item \texttt{add}:
  \begin{equation}
    \mathrm{Fuse}(\mathbf{H}_s,\mathbf{H}_\ell)=\mathbf{H}_s+\mathbf{H}_\ell.
  \end{equation}

  \item \texttt{weighting}:
  The specific output form in this fusion family is selected by the conditional variable $x_{23}$.

  \item \texttt{gating}:
  \begin{equation}
    \mathbf{G}=\sigma\!\left(\mathbf{W}_g[\mathbf{H}_s\|\mathbf{H}_\ell]+\mathbf{b}_g\right),\qquad
    \mathrm{Fuse}(\mathbf{H}_s,\mathbf{H}_\ell)=\mathbf{G}\odot\mathbf{H}_s+(1-\mathbf{G})\odot\mathbf{H}_\ell.
  \end{equation}

  \item \texttt{attention} (cross-attention form):
  Here we use single-head cross-attention and fix $d=C_f$, where $C_f$ equals the decoded value of $x_9$ in Table~\ref{tab:decision-space}.
  Let $\mathbf{H}_s,\mathbf{H}_\ell\in\mathbb{R}^{L_p\times C_f}$, and let $\mathbf{W}_Q,\mathbf{W}_K,\mathbf{W}_V\in\mathbb{R}^{C_f\times C_f}$.
  \begin{equation}
    \mathbf{Q}=\mathbf{H}_s\mathbf{W}_Q,\quad
    \mathbf{K}=\mathbf{H}_\ell\mathbf{W}_K,\quad
    \mathbf{V}=\mathbf{H}_\ell\mathbf{W}_V,\quad
    \mathrm{Fuse}(\mathbf{H}_s,\mathbf{H}_\ell)=\mathrm{softmax}\!\left(\frac{\mathbf{Q}\mathbf{K}^\top}{\sqrt{d}}\right)\mathbf{V}.
  \end{equation}

  \item \texttt{cross\_mapping}:
  The specific combination mode in this fusion family is selected by the conditional variable $x_{24}$.

\end{enumerate}

\paragraph{$x_{23}$: Weighting mode (active if $x_{22}=\texttt{weighting}$).}
Active only when $x_{22}=\texttt{weighting}$, we first compute a weight $\alpha\in[0,1]$ and then output according to $x_{23}$.
Here $\mathrm{GAP}(\cdot)$ denotes global average pooling over the time dimension.
\begin{equation}
  \alpha=\sigma\!\left(w^\top \mathrm{GAP}([\mathbf{H}_s\|\mathbf{H}_\ell])\right).
\end{equation}

\begin{enumerate}
  \renewcommand{\labelenumi}{(\arabic{enumi})}
  \item \texttt{add}:
  \begin{equation}
    \mathrm{Fuse}(\mathbf{H}_s,\mathbf{H}_\ell)=\alpha\,\mathbf{H}_s+(1-\alpha)\,\mathbf{H}_\ell.
  \end{equation}
  \item \texttt{concat}:
  \begin{equation}
    \mathrm{Fuse}(\mathbf{H}_s,\mathbf{H}_\ell)=\left[\alpha\,\mathbf{H}_s\ \|\ (1-\alpha)\,\mathbf{H}_\ell\right].
  \end{equation}
\end{enumerate}

\paragraph{$x_{24}$: Cross-mapping mode (active if $x_{22}=\texttt{cross\_mapping}$).}
Active only when $x_{22}=\texttt{cross\_mapping}$, we first perform cross-branch mappings and then combine according to $x_{24}$.
\begin{equation}
  \hat{\mathbf{H}}_s=\phi_{s}(\mathbf{H}_\ell),\qquad \hat{\mathbf{H}}_\ell=\phi_{\ell}(\mathbf{H}_s),
\end{equation}
where $\phi_s,\phi_\ell$ are learnable pointwise linear mappings (e.g., $1\times1$ convolutions or linear layers).

\begin{enumerate}
  \renewcommand{\labelenumi}{(\arabic{enumi})}
  \item \texttt{add}:
  \begin{equation}
    \mathrm{Fuse}(\mathbf{H}_s,\mathbf{H}_\ell)=\left(\mathbf{H}_s+\hat{\mathbf{H}}_s\right)+\left(\mathbf{H}_\ell+\hat{\mathbf{H}}_\ell\right).
  \end{equation}
  \item \texttt{concat}:
  \begin{equation}
    \mathrm{Fuse}(\mathbf{H}_s,\mathbf{H}_\ell)=\left[\mathbf{H}_s+\hat{\mathbf{H}}_s\ \|\ \mathbf{H}_\ell+\hat{\mathbf{H}}_\ell\right].
  \end{equation}
  \item \texttt{gated}:
  \begin{equation}
    \mathbf{G}=\sigma\!\left(\mathbf{W}_g[\mathbf{H}_s\|\mathbf{H}_\ell]+\mathbf{b}_g\right),
    \quad
    \mathrm{Fuse}(\mathbf{H}_s,\mathbf{H}_\ell)=\mathbf{G}\odot(\mathbf{H}_s+\hat{\mathbf{H}}_s)+(1-\mathbf{G})\odot(\mathbf{H}_\ell+\hat{\mathbf{H}}_\ell).
  \end{equation}
\end{enumerate}

\subsection{Evolutionary Optimization Metrics}
\label{appendix:evo_metrics}
Let the reference Pareto front be $P^\ast=\{z_1, z_2, \ldots, z_m\}$ and the non-dominated solution set obtained by an algorithm be $A=\{a_1, a_2, \ldots, a_n\}$.

\paragraph{Inverted Generational Distance (IGD)}
IGD measures the average distance from each point on the reference front to its nearest point in $A$:
\begin{equation}
  \mathrm{IGD}(A, P^\ast)
  =
  \frac{1}{|P^\ast|}
  \sum_{z \in P^\ast} \min_{a \in A} \|z-a\|_2 .
\end{equation}

\paragraph{Hypervolume (HV)}
Given a reference point $r$, HV measures the Lebesgue measure of the objective-space region dominated by $A$ and bounded by $r$:
\begin{equation}
  \mathrm{HV}(A)
  =
  \mathrm{Vol}\!\left(
    \bigcup_{a \in A} [a, r]
  \right).
\end{equation}

\subsection{Model Prediction Metrics}
\label{appendix:model_metrics}
Assume the test set contains $N$ samples. For the $n$-th sample and the $k$-th target ($k=1,\dots,K$, $n=1,\dots,N$), the ground truth and prediction are denoted by $y_{k,n}$ and $\hat{y}_{k,n}$, respectively. Let $\bar{y}_k=\frac{1}{N}\sum_{n=1}^{N}y_{k,n}$.

\paragraph{Per-target metrics}
The mean squared error (MSE), mean absolute error (MAE), and mean absolute percentage error (MAPE) for target $k$ are defined as
\begin{equation}
  \mathrm{MSE}_k
  =
  \frac{1}{N}
  \sum_{n=1}^{N}
  \bigl(y_{k,n} - \hat{y}_{k,n}\bigr)^2,
\end{equation}
\begin{equation}
  \mathrm{MAE}_k
  =
  \frac{1}{N}
  \sum_{n=1}^{N}
  \bigl|y_{k,n} - \hat{y}_{k,n}\bigr|,
\end{equation}
\begin{equation}
  \mathrm{MAPE}_k
  =
  \frac{100}{N}
  \sum_{n=1}^{N}
  \left|
    \frac{y_{k,n} - \hat{y}_{k,n}}{y_{k,n} + \epsilon}
  \right|,
\end{equation}
where $\epsilon$ is a small constant to avoid division by zero in degenerate cases.

\paragraph{Across-target averaged metrics}
The mean MAPE across $K$ targets is
\begin{equation}
  \mathrm{MAPE}_{\mathrm{mean}}
  =
  \frac{1}{K}
  \sum_{k=1}^{K}
  \mathrm{MAPE}_k.
\end{equation}

\paragraph{Normalized metrics (NMSE and NMAE)}
Let $\sigma_k$ be the standard deviation of the ground-truth values for target $k$ (computed on the evaluation set or on a fixed reference set, consistent across methods). The overall normalized metrics are defined by normalizing each target and then averaging over targets:
\begin{equation}
\label{eq:mse}
  \mathrm{NMSE}
  =
  \frac{1}{K}
  \sum_{k=1}^{K}
  \frac{\mathrm{MSE}_k}{\sigma_k^2 + \epsilon},
  \qquad
  \mathrm{NMAE}
  =
  \frac{1}{K}
  \sum_{k=1}^{K}
  \frac{\mathrm{MAE}_k}{\sigma_k + \epsilon}.
\end{equation}
In this work, $K=5$.

\subsection{Experimental Settings and Hardware}
\label{app:exp-settings}

Software implementation: all code is implemented and debugged in PyCharm (Python/PyTorch).

Hardware: since one FE on the real sintering task corresponds to a full train-and-validate run, we perform parallel evaluations on a server equipped with two RTX 3090 GPUs; model development and debugging are conducted on a local Apple M4 Pro MacBook Pro, which can remotely schedule the server resources under the same experimental pipeline.

Algorithm settings (PHMOEA and evolutionary baselines): on the real sintering task, we use a population size $N=50$ and a maximum generation $T_{\max}=30$, resulting in a maximum budget of $N\!\cdot\!T_{\max}=1500$ function evaluations (FEs).
All evolutionary methods share the same budget and termination criterion on the real task, and we enable the same early-stopping mechanism with a window size $W=8$ generations.
On synthetic benchmarks, to visualize complete generation-wise convergence trajectories, we disable early stopping and set $N=100$ and $T_{\max}=100$.
PHMOEA uses the following fixed hyper-parameters: crossover probability $p_c=0.8$, mutation probability $p_m=0.2$, initial bin number $K_i=6$ for continuous variables, and deduplication resampling limit $n_{\mathrm{trial}}=50$.
Player tracking uses $q=0.3$, $p=0.2$, cold bonus $o=0.15$, and cross-pool mutation probability $e=0.1$.
We use the stage factor $\phi_t=t/T_{\max}$ and split the search into three stages using $\kappa_1$ and $\kappa_2$: early if $\phi_t<\kappa_1$, middle if $\kappa_1\le \phi_t<\kappa_2$, and late if $\phi_t\ge \kappa_2$.
We set $(\kappa_1,\kappa_2)=(0.3,0.6)$ for the real sintering task and $(\kappa_1,\kappa_2)=(0.2,0.4)$ for synthetic benchmarks.
Stage ratios are controlled by $\operatorname{StageRatio}(\phi_t)$ using a fixed three-stage schedule: in the early stage, $(\rho_{\mathrm{par}},\rho_{\mathrm{hot}},\rho_{\mathrm{nh}})=(0.8,0.1,0.1)$; in the middle stage, $(0.6,0.2,0.2)$; and in the late stage, $(0.5,0.3,0.2)$.
For continuous variables, we use SBX crossover and polynomial mutation (PM) with distribution indices $\eta_c=15$ and $\eta_m=20$, and apply polynomial mutation with a per-dimension probability $p_m^{\mathrm{poly}}=1/D$, where $D$ is the encoding dimension.
We cap the number of mutated dimensions in a single mutation to $m_{\max}=\min(6, D)$ to avoid overly aggressive perturbations.
For Eq.(8), we slightly bias toward the error objective on the real task by setting $w=0.7$, while using balanced weights $w=0.5$ on synthetic benchmarks; the remaining coefficients are fixed to $\lambda=0.2$ and $\gamma=0.05$ in all experiments.
Unless otherwise stated, each evolutionary algorithm is run for 20 independent trials and we report mean$\pm$std, with statistical significance tested by the Wilcoxon rank-sum test at $p<0.05$.

Model settings (MS--BCNN as a PHMOEA-selected configuration): PHMOEA outputs a Pareto set of non-dominated solutions; to showcase the instantiated model and conduct downstream forecasting-performance comparisons, we select one solution as an exemplar configuration $x^{\mathrm{sel}}$ from the Pareto set and keep it fixed across all forecasting comparisons. To avoid ambiguity, we record its instantiated decision-variable values following Table~\ref{tab:decision-space} (candidate sets in Table~\ref{tab:decision-space}; operator/loss definitions in Appendix~\ref{app:resampling-ops}):

\begin{align*}
&x_1=\texttt{hybrid},\;
x_2=\texttt{avg}\;(\text{inactive if }x_1\neq\texttt{pool}),\;
x_3=12,\;
x_4=32,\;
x_5=\texttt{BatchNorm},\\
&x_6=16,\;
x_7=16,\;
x_8=32,\;
x_9=64,\;
x_{10}=(3,5,7),\;
x_{11}=(9,11,13),\\
&x_{12}=\texttt{Tanh},\;
x_{13}=0.1,\;
x_{14}=10^{-3},\;
x_{15}=5.39\times 10^{-5},\;
x_{16}=\texttt{on},\\
&x_{17}=\texttt{warmup cosine}\;(\text{active if }x_{16}=\texttt{on}),\;
x_{18}=\texttt{AdaptiveCombined},\;
x_{19}=(\texttt{MAE},\texttt{LogCosh})\;(\text{active if }x_{18}\in\{\texttt{Combined},\texttt{AdaptiveCombined}\}),\\
&x_{20}=(0.7,0.3)\;(\text{active if }x_{18}=\texttt{AdaptiveCombined}),\;
x_{21}=0.001\;(\text{active if }x_{18}=\texttt{AdaptiveCombined}),\\
&x_{22}=\texttt{concat},\;
x_{23}=\texttt{--}\;(\text{inactive if }x_{22}\neq\texttt{weighting}),\;
x_{24}=\texttt{--}\;(\text{inactive if }x_{22}\neq\texttt{cross\_mapping}).
\end{align*}

\section{Full Results and Ablations}
\label{app:ablation}

\subsection{Results of Evolutionary Algorithms}

\begin{table}[H]
  \centering
  \caption{Per-benchmark evolutionary results on H-DTLZ2/7 at the final generation ($g{=}100$).
  Best IGD (lower) and best HV (higher) per benchmark are highlighted in gray. Values are reported with 7 decimal places.}
  \label{tab:evo_hdtlz_27}
  \small
  \setlength{\tabcolsep}{6pt}
  \renewcommand{\arraystretch}{1.05}
  \begin{tabular}{llcc}
    \toprule
    Benchmark & Method & IGD $\downarrow$ & HV $\uparrow$ \\
    \midrule
    \multirow{5}{*}{H-DTLZ2}
      & PHMOEA   & 0.0051725$\pm$0.00008 & 0.4188035$\pm$0.00024 \\
      & MOEA/D   & 0.0043681$\pm$0.00011 & 0.4192696$\pm$0.00041 \\
      & NSGA-II  & 0.0049416$\pm$0.00038 & 0.4194225$\pm$0.00062 \\
      & NSGA-III & \cellcolor{gray!25}0.0041721$\pm$0.00002 & 0.4197837$\pm$0.00015 \\
      & SMSEMOA  & 0.0051574$\pm$0.00012 & \cellcolor{gray!25}0.4208402$\pm$0.00004 \\
    \midrule
    \multirow{5}{*}{H-DTLZ7}
      & PHMOEA   & 0.0033302$\pm$0.00013 & 0.5566254$\pm$0.00001 \\
      & MOEA/D   & 0.0033004$\pm$0.00016 & 0.5566116$\pm$0.00001 \\
      & NSGA-II  & 0.0034201$\pm$0.00013 & 0.5564066$\pm$0.00008 \\
      & NSGA-III & 0.0032780$\pm$0.00001 & 0.5566078$\pm$0.00004 \\
      & SMSEMOA  & \cellcolor{gray!25}0.0025751$\pm$0.000015 & \cellcolor{gray!25}0.5572028$\pm$0.00009 \\
    \bottomrule
  \end{tabular}
\end{table}

\begin{figure}[H]
\centering
\includegraphics[width=\linewidth]{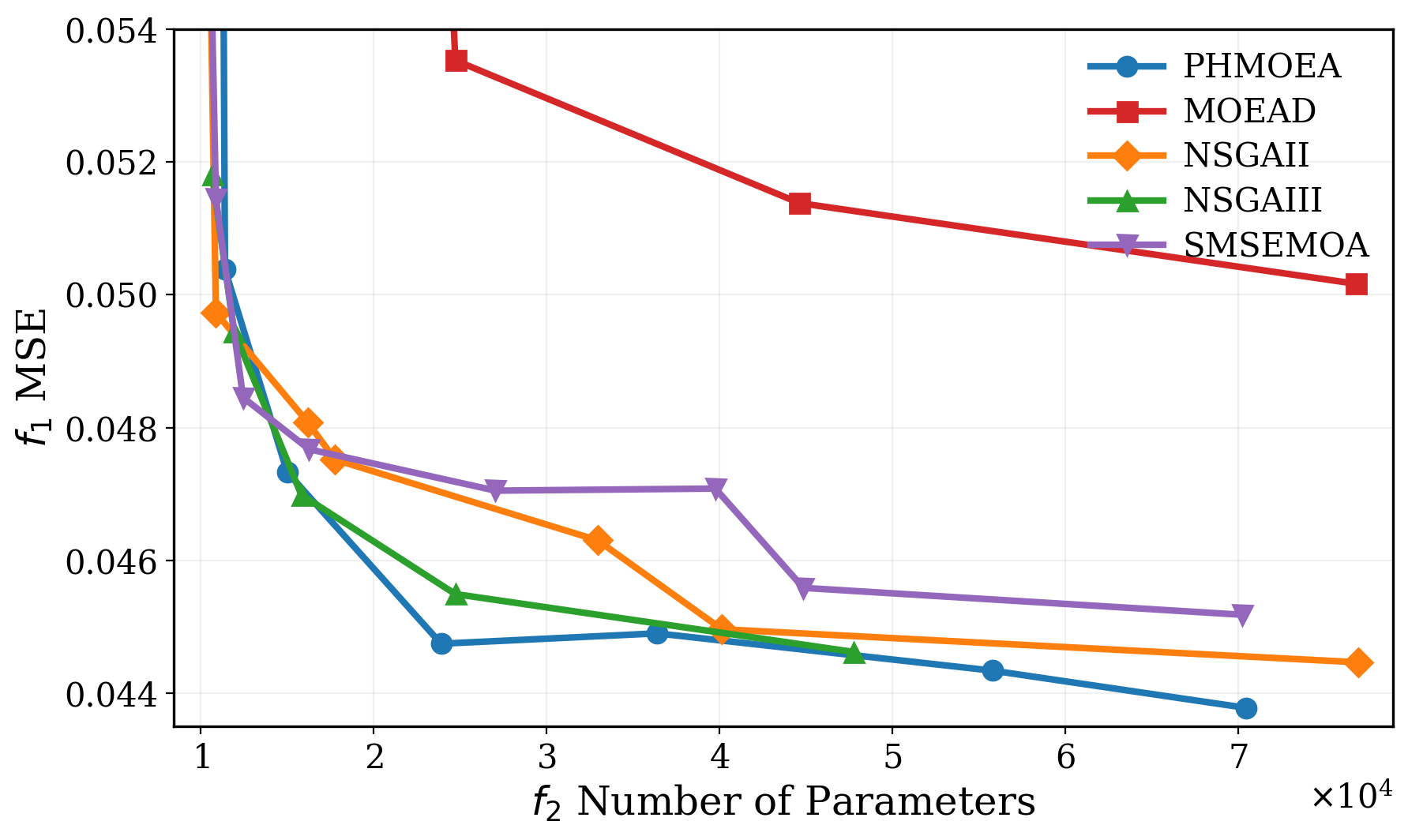}
\caption{Comparison of Pareto fronts achieved by different evolutionary algorithms on the real-world sintering task in the MSE--parameter-count objective space.}
\label{fig:evo_pareto_app}
\end{figure}

\subsection{Results of  Models}

\begin{table}[H]
  \centering
  
 \caption{Per-target forecasting performance on five sintering targets under the shuffled (approximately i.i.d.) evaluation setting (mean$\pm$std). For each target and metric, the best result is highlighted in gray.}
  \label{tab:static_full}
  \begin{tabular}{llccc}
    \toprule
    Model & Target & MSE / NMSE & MAE / NMAE & MAPE (\%) \\
    \midrule
    \multirow{6}{*}{MS-BCNN}
      & TFe      & \cellcolor{gray!50}{$0.066887 \pm 0.004481$} & \cellcolor{gray!50}{$0.199272 \pm 0.006370$} & \cellcolor{gray!50}{$0.358438 \pm 0.011667$} \\
      & FeO      & $0.130500 \pm 0.005977$ & \cellcolor{gray!50}{$0.248728 \pm 0.007786$} & \cellcolor{gray!50}{$2.762939 \pm 0.082877$} \\
      & SiO$_2$  & \cellcolor{gray!50}{$0.004147 \pm 0.000860$} & \cellcolor{gray!50}{$0.047211 \pm 0.003289$} & \cellcolor{gray!50}{$0.834882 \pm 0.055914$} \\
      & CaO      & \cellcolor{gray!50}{$0.034386 \pm 0.002732$} & \cellcolor{gray!50}{$0.137910 \pm 0.004608$} & \cellcolor{gray!50}{$1.140071 \pm 0.037974$} \\
      & Basicity & \cellcolor{gray!50}{$0.000561 \pm 0.000020$} & \cellcolor{gray!50}{$0.017901 \pm 0.000419$} & \cellcolor{gray!50}{$0.840955 \pm 0.019404$} \\
      & Overall  & \cellcolor{gray!50}{$0.267880 \pm 0.217387$} & \cellcolor{gray!50}{$0.347816 \pm 0.138165$} & \cellcolor{gray!50}{$1.187457 \pm 0.041567$} \\
    \midrule
    \multirow{6}{*}{OB-ISSID}
      & TFe      & $0.087376 \pm 0.012018$ & $0.221529 \pm 0.013426$  & $0.397920 \pm 0.023850$  \\
      & FeO      & $0.124807 \pm 0.009778$ & $0.259250 \pm 0.008807$ & $2.904489 \pm 0.101035$ \\
      & SiO$_2$  & $0.006618 \pm 0.001655$ & $0.056789 \pm 0.005482$ & $1.007978 \pm 0.101921$  \\
      & CaO      & $0.047292 \pm 0.008086$ & $0.156763 \pm 0.009696$ & $1.302444 \pm 0.085195$ \\
      & Basicity & $0.000690 \pm 0.000050$ & $0.019374 \pm 0.000459$ & $0.912564 \pm 0.021654$ \\
      & Overall  & $0.304435 \pm 0.200582$ & $0.380942 \pm 0.135334$ & $1.305079 \pm 0.066731$ \\
    \midrule
    \multirow{6}{*}{Ventingformer}
      & TFe      & $0.078146 \pm 0.012676$ & $0.209559 \pm 0.015034$  & $0.376626 \pm 0.027000$  \\
      & FeO      & \cellcolor{gray!50}{$0.121521 \pm 0.013489$} & $0.252890 \pm 0.011607$ & $2.829842 \pm 0.137632$ \\
      & SiO$_2$  & $0.004980 \pm 0.001040$ & $0.052148 \pm 0.004801$ & $0.924745 \pm 0.085734$ \\
      & CaO      & $0.041254 \pm 0.006052$ & $0.149523 \pm 0.007049$ & $1.240762 \pm 0.060862$ \\
      & Basicity & $0.000623 \pm 0.000035$ & $0.018812 \pm 0.000359$ & $0.885587 \pm 0.016939$ \\
      & Overall  & $0.279285 \pm 0.198880$ & $0.365572 \pm 0.136393$ & $1.251512 \pm 0.065633$ \\
    \midrule
    \multirow{6}{*}{Transformer}
      & TFe      & $0.179614 \pm 0.051713$ & $0.306316 \pm 0.043242$  & $0.548945 \pm 0.076513$  \\
      & FeO      & $0.146094 \pm 0.013018$ & $0.281115 \pm 0.012444$ & $3.140294 \pm 0.141931$ \\
      & SiO$_2$  & $0.018645 \pm 0.006535$ & $0.096401 \pm 0.017427$ & $1.725985 \pm 0.322216$  \\
      & CaO      & $0.089728 \pm 0.021513$ & $0.213125 \pm 0.026726$ & $1.784003 \pm 0.234445$ \\
      & Basicity & $0.000906 \pm 0.000114$ & $0.022628 \pm 0.001780$ & $1.067092 \pm 0.084132$ \\
      & Overall  & $0.456411 \pm 0.176579$ & $0.482932 \pm 0.102764$ & $1.653264 \pm 0.171847$ \\
    \midrule
    \multirow{6}{*}{LSTM}
      & TFe      & $0.121748 \pm 0.021304$ & $0.247516 \pm 0.015248$ & $0.444744 \pm 0.026978$ \\
      & FeO      & $0.143653 \pm 0.012786$ & $0.278965 \pm 0.010682$ & $3.122390 \pm 0.109783$ \\
      & SiO$_2$  & $0.010181 \pm 0.002500$ & $0.066771 \pm 0.006445$ & $1.185119 \pm 0.120935$ \\
      & CaO      & $0.062829 \pm 0.012811$ & $0.170975 \pm 0.010754$ & $1.423044 \pm 0.094781$ \\
      & Basicity & $0.000855 \pm 0.000080$ & $0.021571 \pm 0.001170$ & $1.018066 \pm 0.056480$ \\
      & Overall  & $0.380269 \pm 0.220356$ & $0.421303 \pm 0.141852$ & $1.438673 \pm 0.081791$ \\
    \midrule
    \multirow{6}{*}{GRU-PLS}
      & TFe      & $0.202637 \pm 0.018764$ & $0.354862 \pm 0.017037$  & $0.635633 \pm 0.030851$  \\
      & FeO      & $0.146588 \pm 0.017297$ & $0.281162 \pm 0.011820$ & $3.148090 \pm 0.132091$ \\
      & SiO$_2$  & $0.018845 \pm 0.001616$ & $0.106709 \pm 0.004294$ & $1.898254 \pm 0.075795$  \\
      & CaO      & $0.087245 \pm 0.008711$ & $0.231595 \pm 0.011355$ & $1.927161 \pm 0.094707$ \\
      & Basicity & $0.000894 \pm 0.000073$ & $0.022799 \pm 0.000951$ & $1.075404 \pm 0.046638$ \\
      & Overall  & $0.461843 \pm 0.170664$ & $0.512008 \pm 0.082071$ & $1.736908 \pm 0.076016$ \\
    \bottomrule
  \end{tabular}

  \vspace{2mm}
  {\footnotesize\textit{Note:} The \textit{Overall} row reports NMSE and NMAE (computed by normalizing each target's error using its variance/standard deviation and then averaging across the five targets), while MAPE is macro-averaged across the five targets.}
\end{table}

\begin{figure}[H]
  \centering
  \subfloat[MS-BCNN]{%
    \includegraphics[width=0.31\linewidth]{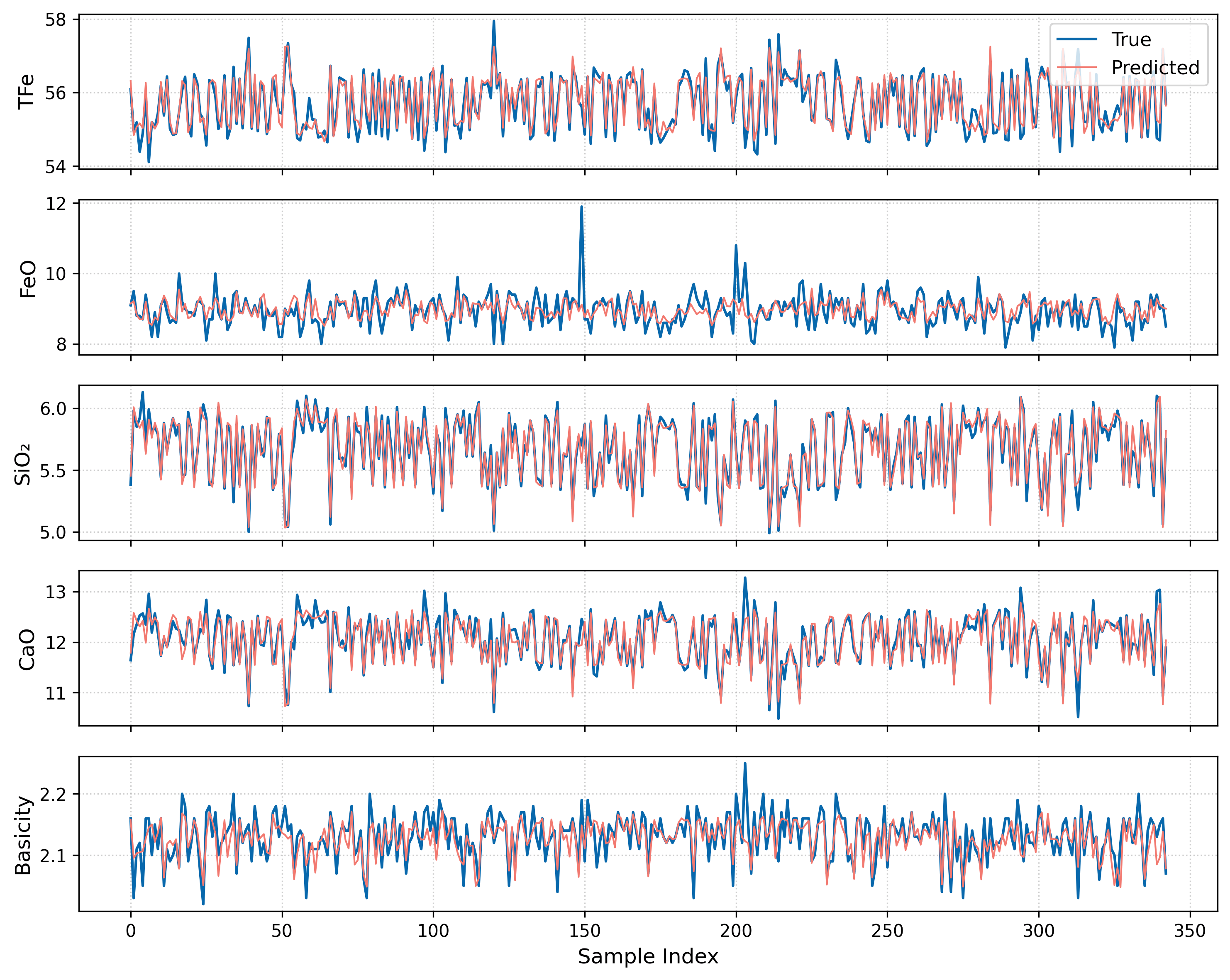}%
  }\hfill
  \subfloat[OB-ISSID]{%
    \includegraphics[width=0.31\linewidth]{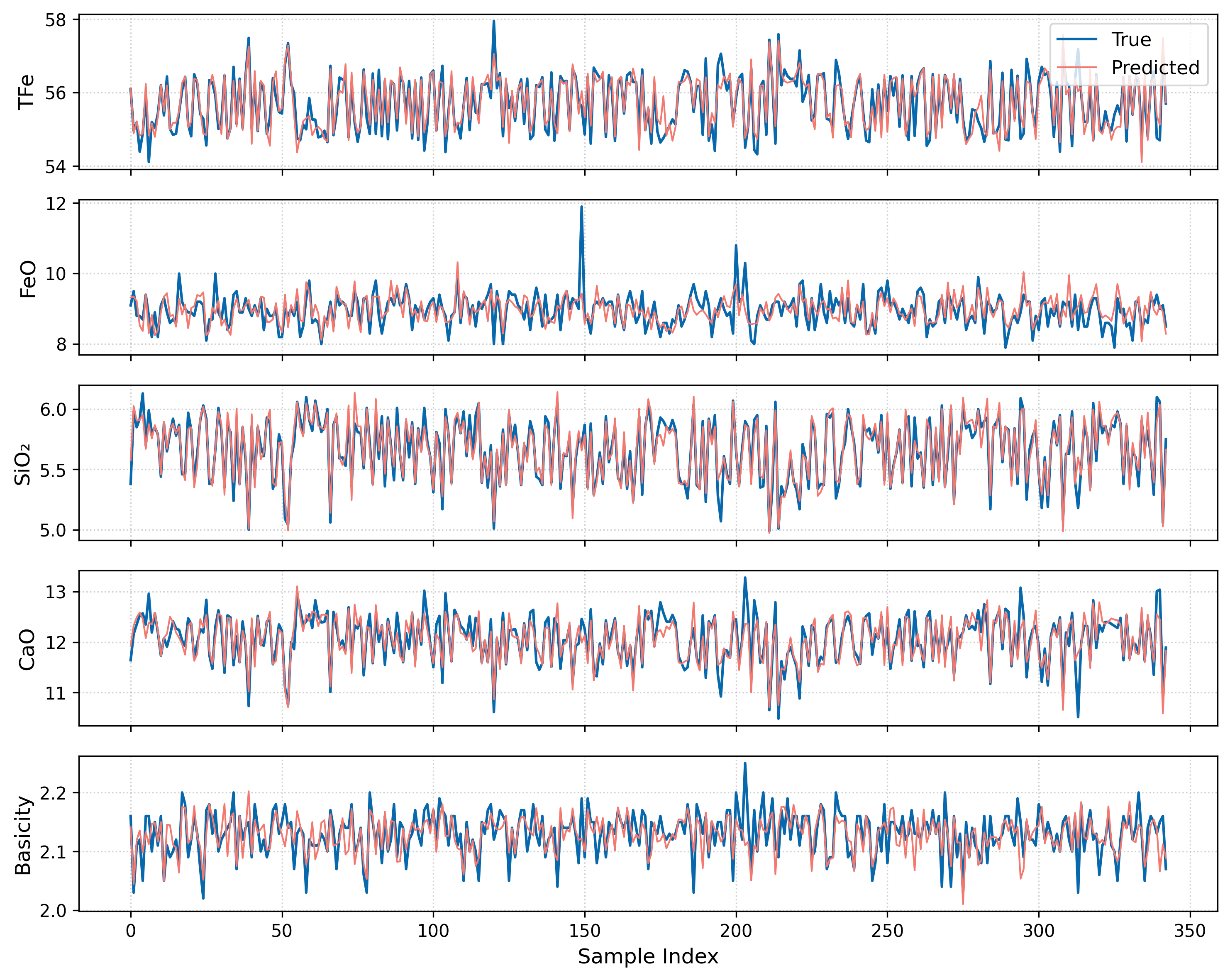}%
  }\hfill
  \subfloat[Ventingformer]{%
    \includegraphics[width=0.31\linewidth]{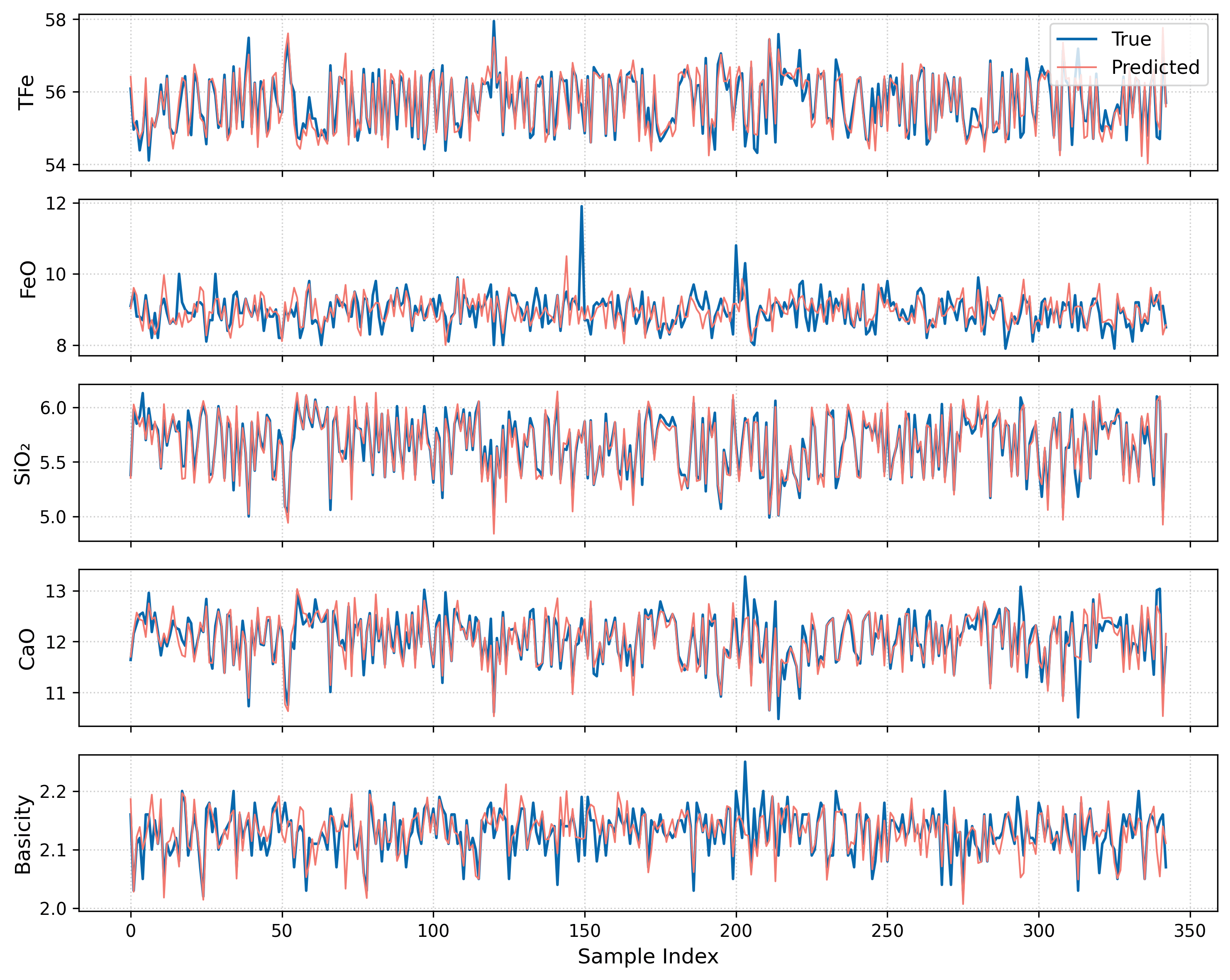}%
  }\\[0.8em]
  \subfloat[Transformer]{%
    \includegraphics[width=0.31\linewidth]{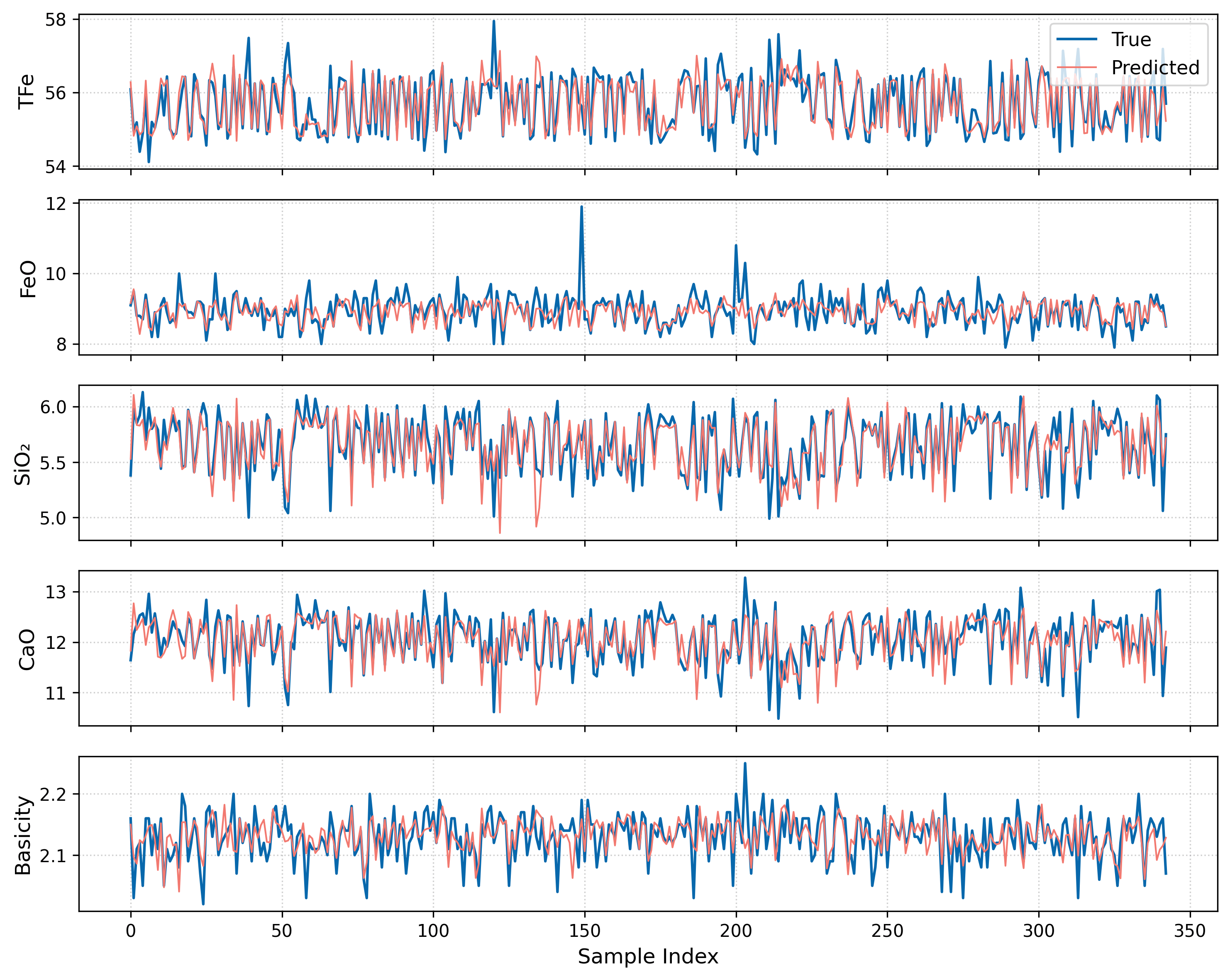}%
  }\hfill
  \subfloat[LSTM]{%
    \includegraphics[width=0.31\linewidth]{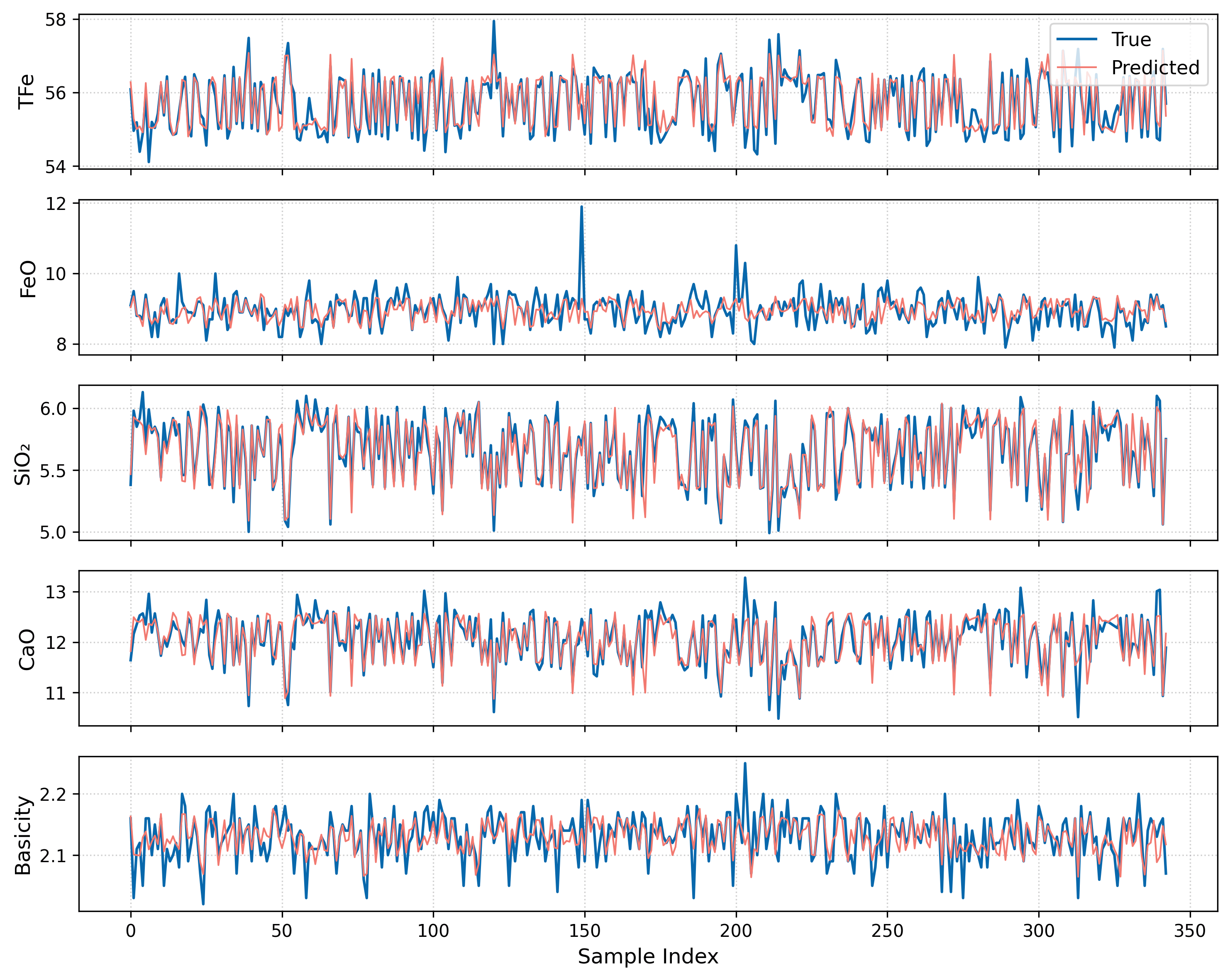}%
  }\hfill
  \subfloat[GRU-PLS]{%
    \includegraphics[width=0.31\linewidth]{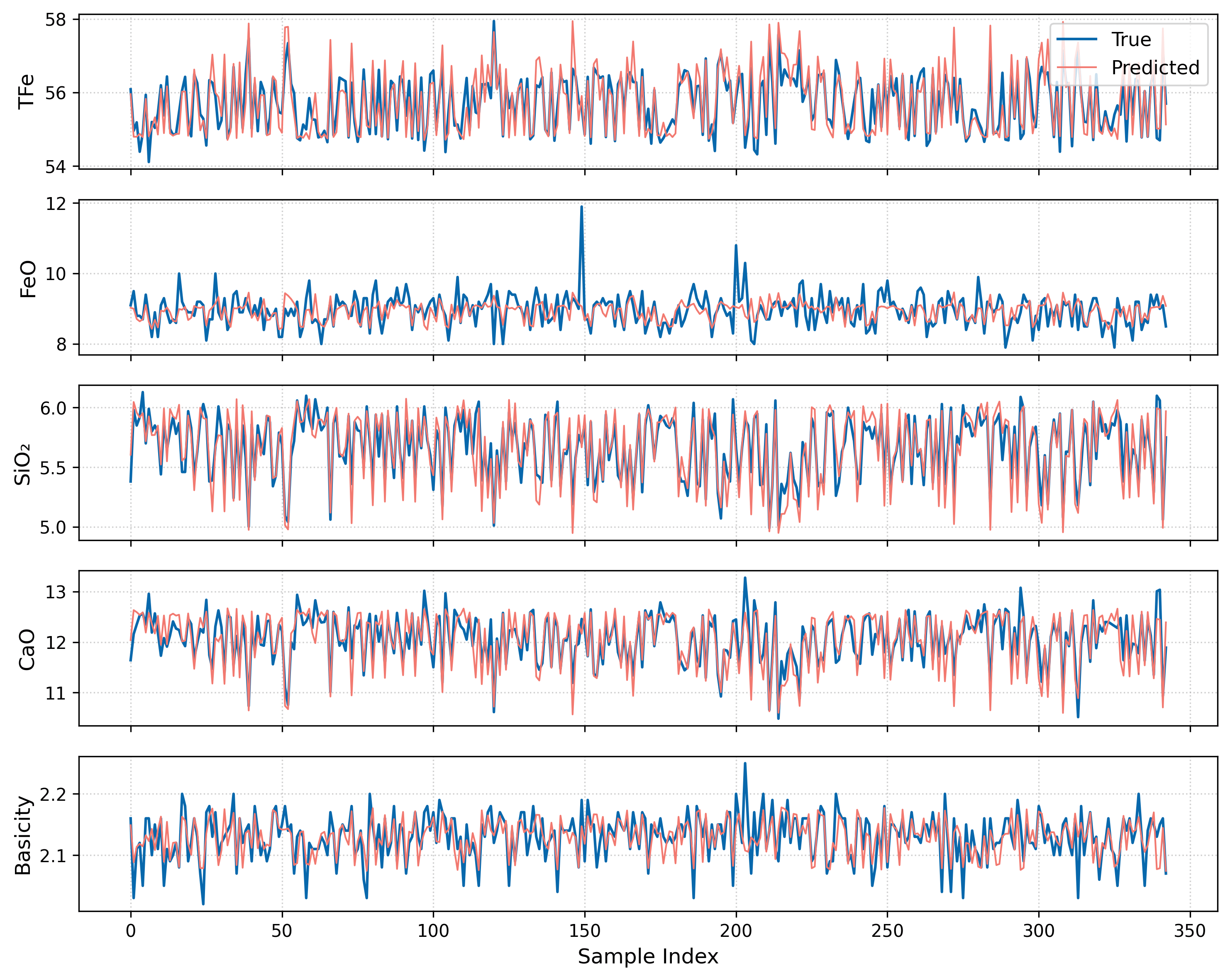}%
  }

  \caption{Qualitative comparison of ground-truth and predicted trajectories for five quality targets under the shuffled (approximately i.i.d.) evaluation setting. Each panel shows the test-set predictions of one model (TFe, FeO, SiO\textsubscript{2}, CaO, and Basicity).}
  \label{fig:static_pred_all}
\end{figure}

\clearpage
\begin{table}[H]
  \centering
  \caption{Per-target forecasting performance on five sintering targets under the chronological (non-shuffled) evaluation setting (mean$\pm$std). For each target and metric, the best result is highlighted in gray.}
  \label{tab:dynamic_full}

  \setlength{\tabcolsep}{3.5pt}
\renewcommand{\arraystretch}{1.05}

  \begin{tabular}{llccc}
    \toprule
    Model & Target & MSE / NMSE & MAE / NMAE & MAPE (\%) \\
    \midrule

    \multirow{6}{*}{MS-BCNN}
      & TFe
      & \cellcolor{gray!50}{$0.387834 \pm 0.141648$}
      & \cellcolor{gray!50}{$0.529125 \pm 0.106408$}
      & \cellcolor{gray!50}{$0.939281 \pm 0.188755$}
       \\
      & FeO
      & \cellcolor{gray!50}{$0.205710 \pm 0.046634$}
      & \cellcolor{gray!50}{$0.344918 \pm 0.047771$}
      & \cellcolor{gray!50}{$3.767097 \pm 0.507709$}
       \\
      & SiO$_2$
      & $0.092268 \pm 0.037270$
      & $0.256975 \pm 0.064096$
      & $4.688675 \pm 1.153858$
       \\
      & CaO
      & $0.340601 \pm 0.136297$
      & $0.479478 \pm 0.110109$
      & $4.048819 \pm 0.919742$
       \\
      & Basicity
      & $0.001920 \pm 0.001217$
      & $0.035376 \pm 0.011207$
      & \cellcolor{gray!50}{$1.639518 \pm 0.517844$}
       \\
      & Overall
      & \cellcolor{gray!50}{$4.445824 \pm 2.193273$}
      & \cellcolor{gray!50}{$1.682359 \pm 0.490118$}
      & $3.016678 \pm 0.657582$
       \\
    \midrule

    \multirow{6}{*}{OB-ISSID}
      & TFe
      & $15.17148 \pm 10.89632$
      & $3.240489 \pm 1.500977$
      & $5.752892 \pm 2.664363$
       \\
      & FeO
      & $7.427538 \pm 11.39653$
      & $1.922083 \pm 1.362099$
      & $21.099157 \pm 14.891946$
       \\
      & SiO$_2$
      & $2.478063 \pm 2.177933$
      & $1.255659 \pm 0.693490$
      & $22.826424 \pm 12.581261$
       \\
      & CaO
      & $2.934882 \pm 1.721921$
      & $1.291617 \pm 0.406435$
      & $10.961348 \pm 3.423734$
       \\
      & Basicity
      & $0.037870 \pm 0.040804$
      & $0.154900 \pm 0.084536$
      & $7.205134 \pm 3.930654$
       \\
      & Overall
      & $119.6787 \pm 101.7280$
      & $8.002796 \pm 3.658696$
      & $13.568991 \pm 7.498392$
       \\
    \midrule

    \multirow{6}{*}{Ventingformer}
      & TFe
      & $0.618011 \pm 0.430214$
      & $0.677437 \pm 0.165248$
      & $1.201339 \pm 0.954510$
       \\
      & FeO
      & $0.344389 \pm 0.161157$
      & $0.494081 \pm 0.124562$
      & $5.503169 \pm 1.158960$
       \\
      & SiO$_2$
      & \cellcolor{gray!50}{$0.069925 \pm 0.004973$}
      & \cellcolor{gray!50}{$0.215943 \pm 0.084562$}
      & $3.937254 \pm 1.425162$
       \\
      & CaO
      & \cellcolor{gray!50}{$0.079916 \pm 0.014892$}
      & \cellcolor{gray!50}{$0.218651 \pm 0.013254$}
      & \cellcolor{gray!50}{$1.851269 \pm 0.242563$}
       \\
      & Basicity
      & $0.002504 \pm 0.000254$
      & $0.042743 \pm 0.021445$
      & $1.976784 \pm 0.581347$
       \\
      & Overall
      & $4.788518 \pm 4.111249$
      & $1.697928 \pm 0.746268$
      & \cellcolor{gray!50}{$2.893963 \pm 0.872508$}
       \\
    \midrule

    \multirow{6}{*}{Transformer}
      & TFe
      & $1.201653 \pm 1.119301$
      & $1.048477 \pm 0.325778$
      & $1.817939 \pm 0.822995$
       \\
      & FeO
      & $0.478953 \pm 0.156598$
      & $0.577473 \pm 0.101278$
      & $6.266190 \pm 0.997501$
       \\
      & SiO$_2$
      & $0.072787 \pm 0.005404$
      & $0.240469 \pm 0.009146$
      & $4.150201 \pm 1.267836$
       \\
      & CaO
      & $0.186698 \pm 0.036337$
      & $0.378812 \pm 0.021531$
      & $3.131392 \pm 0.574227$
       \\
      & Basicity
      & \cellcolor{gray!50}{$0.001879 \pm 0.000277$}
      & \cellcolor{gray!50}{$0.034847 \pm 0.002894$}
      & $1.774490 \pm 0.134311$
       \\
      & Overall
      & $7.580927 \pm 8.678255$
      & $2.180247 \pm 1.303618$
      & $3.428042 \pm 0.759374$
       \\
    \midrule

    \multirow{6}{*}{LSTM}
      & TFe
      & $1.543370 \pm 1.162099$
      & $1.239216 \pm 0.756242$
      & $2.220838 \pm 1.784521$
       \\
      & FeO
      & $0.215657 \pm 0.138634$
      & $0.363922 \pm 0.026993$
      & $3.915424 \pm 1.212547$
       \\
      & SiO$_2$
      & $0.265145 \pm 0.062545$
      & $0.384512 \pm 0.164988$
      & \cellcolor{gray!50}{$3.901176 \pm 1.567876$}
       \\
      & CaO
      & $0.140146 \pm 0.016552$
      & $0.348531 \pm 0.112578$
      & $6.413463 \pm 1.696815$
       \\
      & Basicity
      & $0.358738 \pm 0.148035$
      & $0.549547 \pm 0.193562$
      & $4.699463 \pm 1.364844$
       \\
      & Overall
      & $114.7737 \pm 206.4481$
      & $6.333591 \pm 7.544510$
      & $4.227223 \pm 1.525321$
       \\
    \midrule

    \multirow{6}{*}{GRU-PLS}
      & TFe
      & $0.943971 \pm 0.046501$
      & $0.899228 \pm 0.020768$
      & $1.596798 \pm 0.036822$
       \\
      & FeO
      & $0.387848 \pm 0.036230$
      & $0.521086 \pm 0.030214$
      & $5.807001 \pm 0.337101$
       \\
      & SiO$_2$
      & $0.080240 \pm 0.006853$
      & $0.228669 \pm 0.011692$
      & $4.115747 \pm 0.210896$
       \\
      & CaO
      & $0.959481 \pm 0.052035$
      & $0.880215 \pm 0.024368$
      & $7.426723 \pm 0.206384$
       \\
      & Basicity
      & $0.008617 \pm 0.000501$
      & $0.088803 \pm 0.002844$
      & $4.118412 \pm 0.132350$
       \\
      & Overall
      & $10.71403 \pm 6.279894$
      & $2.795390 \pm 1.053220$
      & $4.612936 \pm 0.184711$
       \\
    \bottomrule
  \end{tabular}

  \vspace{1mm}
  {\scriptsize\textit{Note:} For each target row, we report MSE and MAE (with MAPE). The \textit{Overall} row reports NMSE and NMAE, computed by normalizing each target error (e.g., by its variance/standard deviation) and then macro-averaging over the five targets; MAPE is also macro-averaged.}
\end{table}

\begin{figure}[H]
  \centering

  \subfloat[MS-BCNN]{%
    \includegraphics[width=0.3\linewidth]{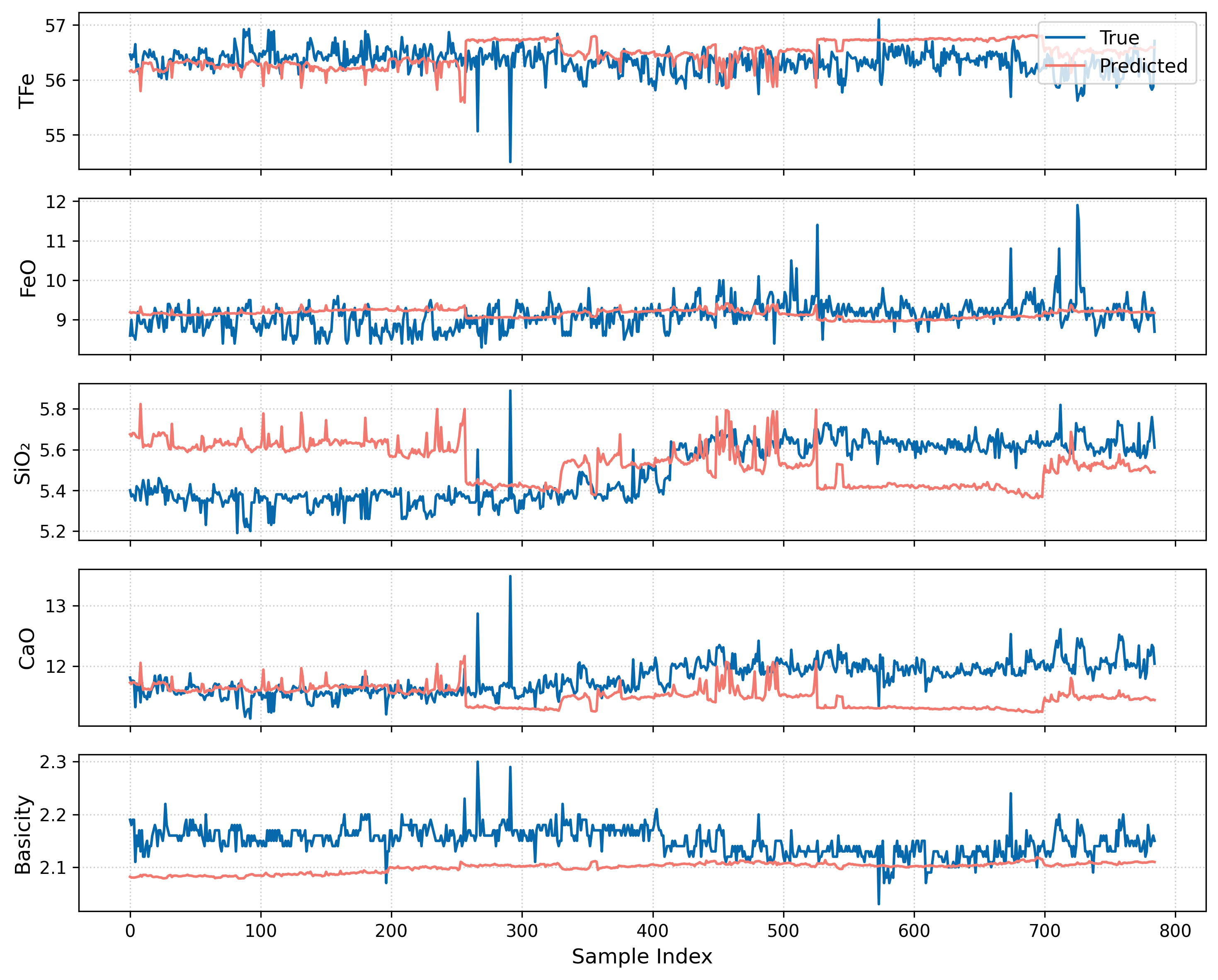}
  }\hfill
  \subfloat[OB-ISSID]{%
    \includegraphics[width=0.3\linewidth]{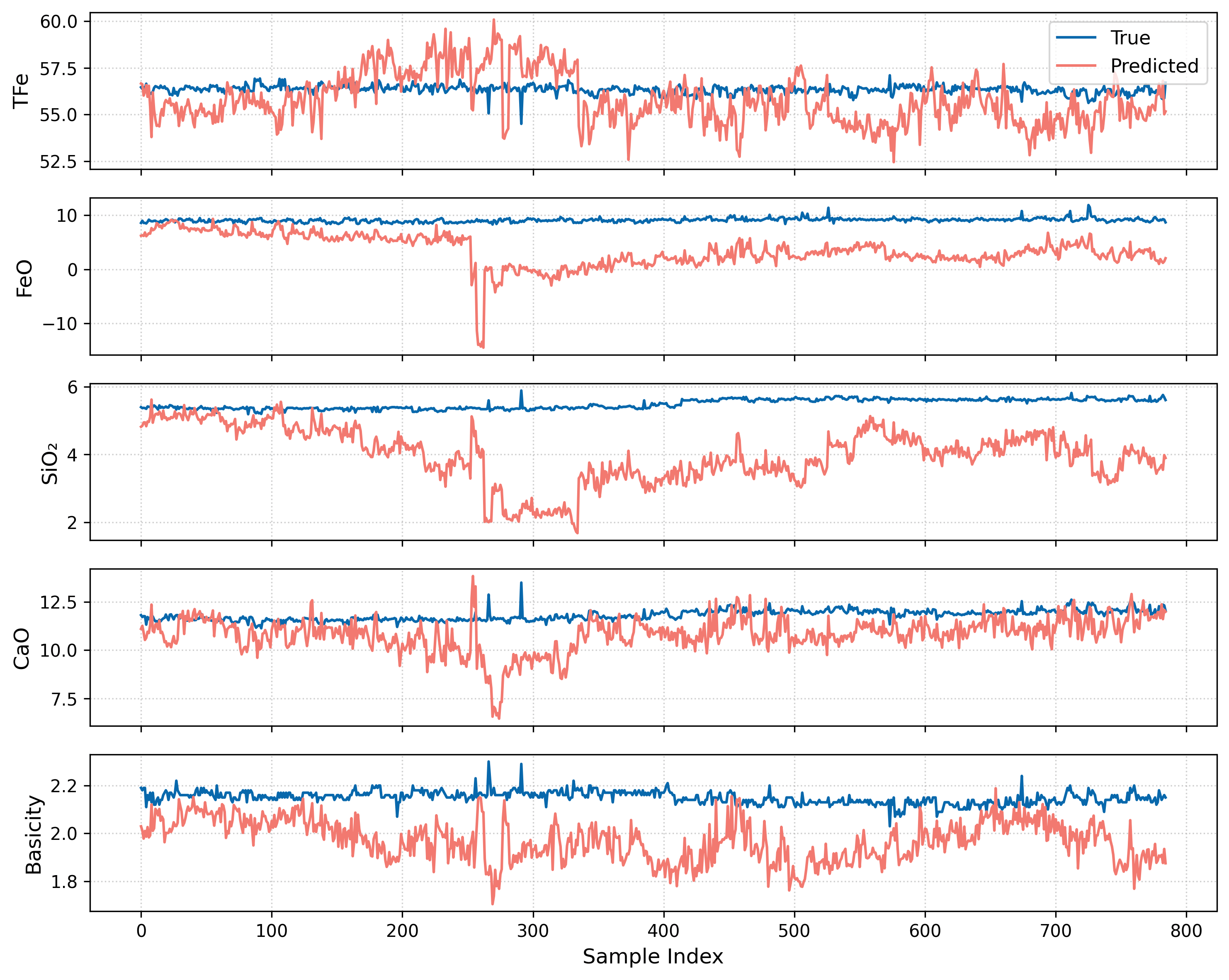}
  }\hfill
  \subfloat[Ventingformer]{%
    \includegraphics[width=0.3\linewidth]{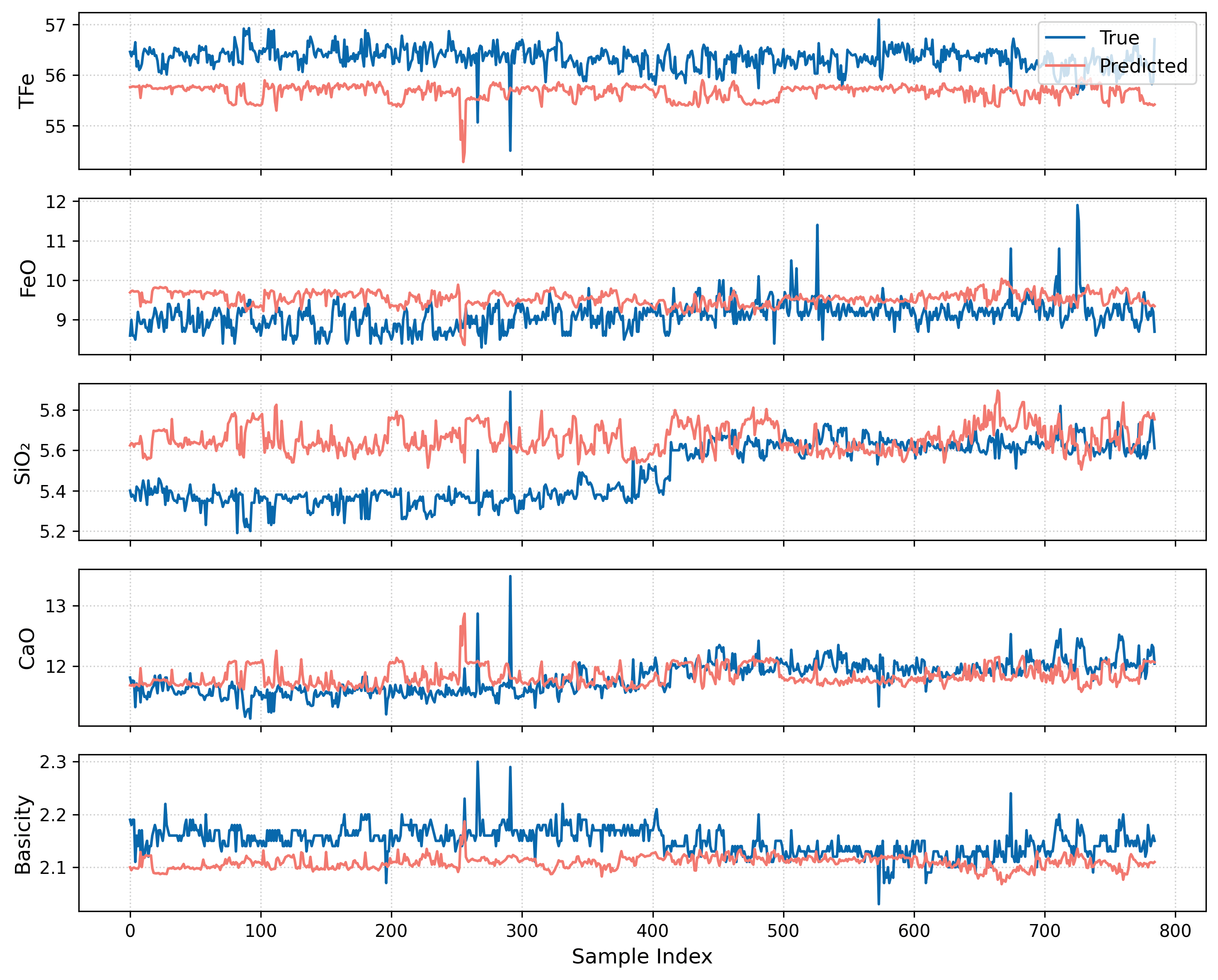}
  }
  \vspace{1em}

  \subfloat[LSTM]{%
    \includegraphics[width=0.3\linewidth]{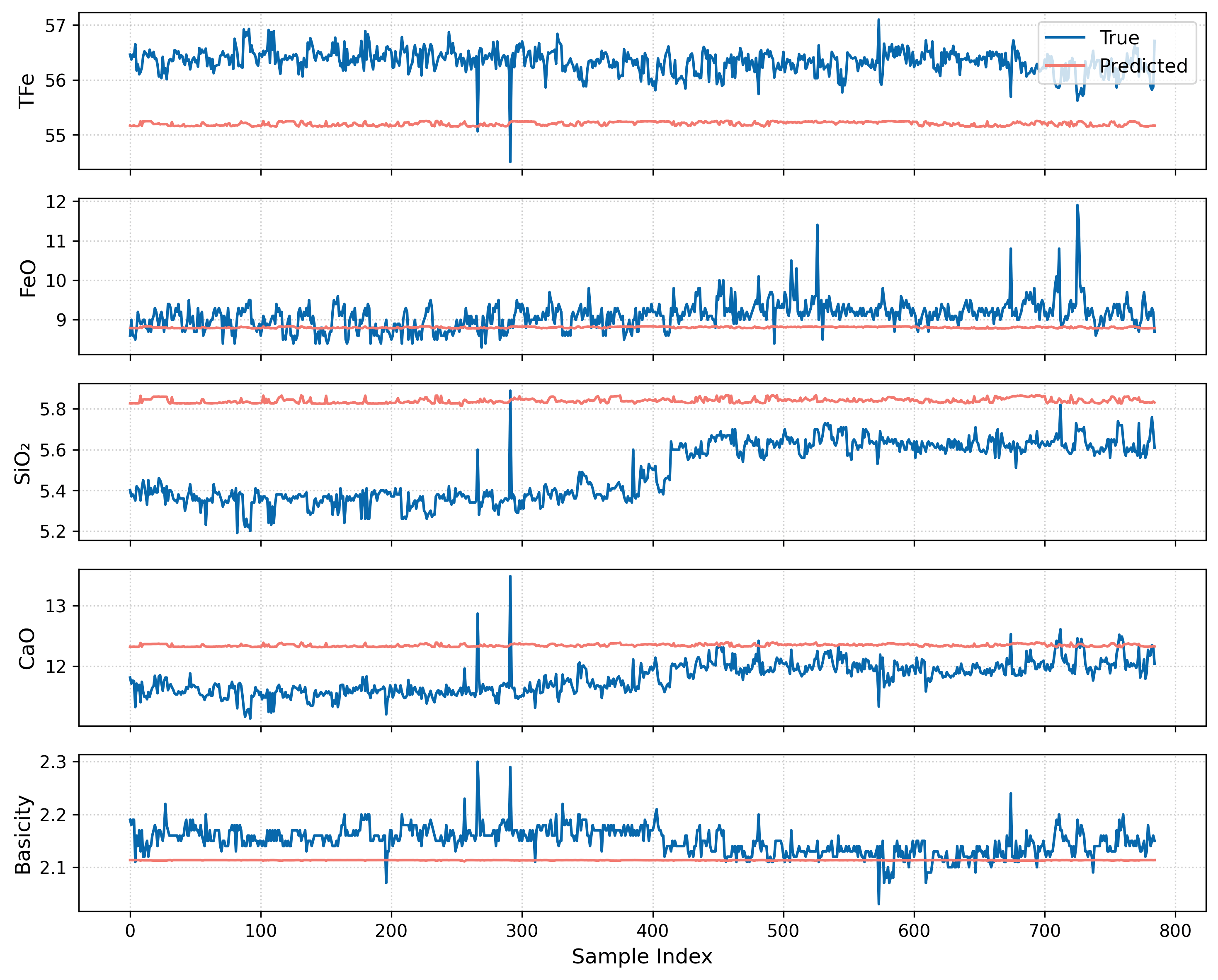}
  }\hfill
  \subfloat[GRU-PLS]{%
    \includegraphics[width=0.3\linewidth]{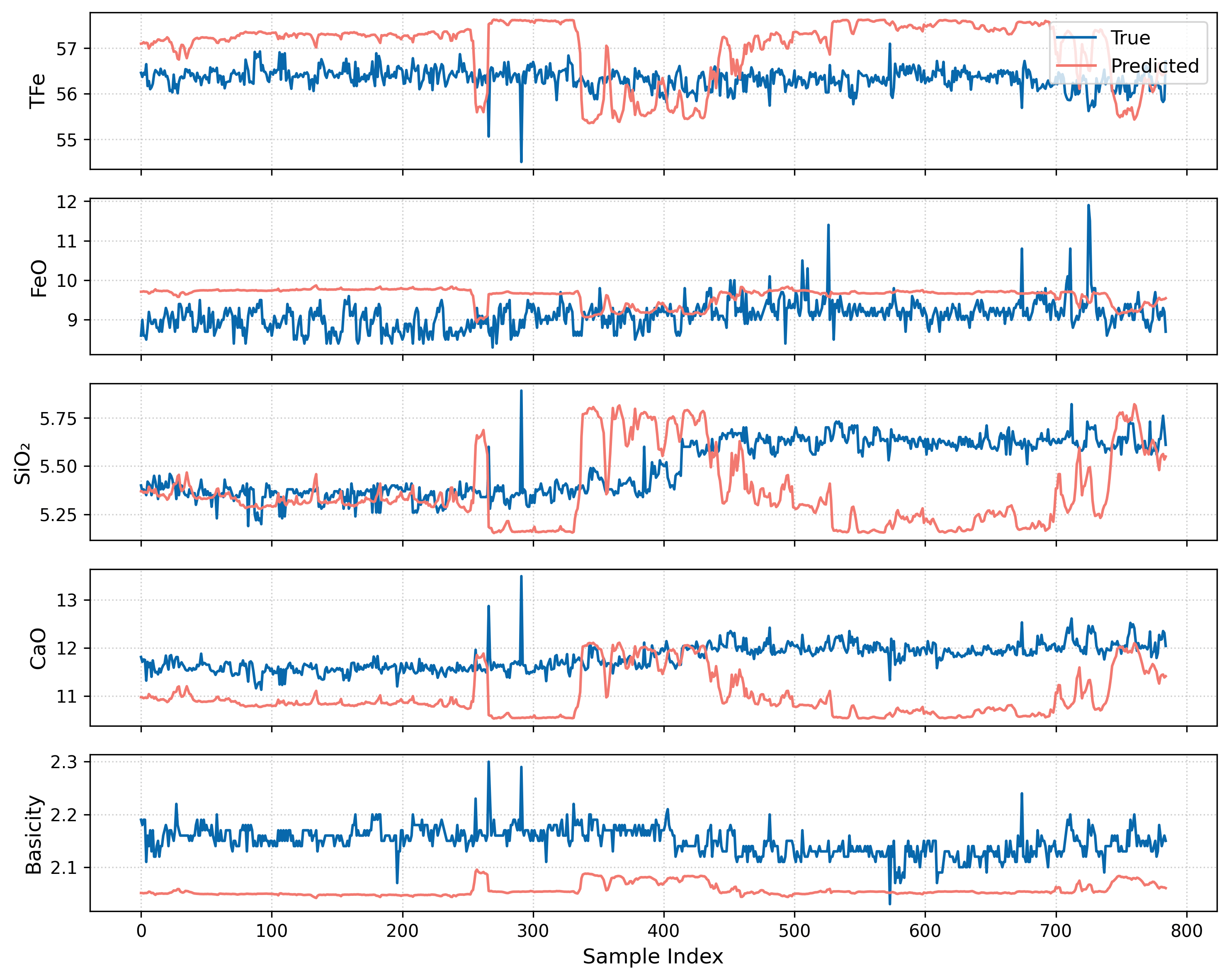}
  }\hfill
  \subfloat[Transformer]{%
    \includegraphics[width=0.3\linewidth]{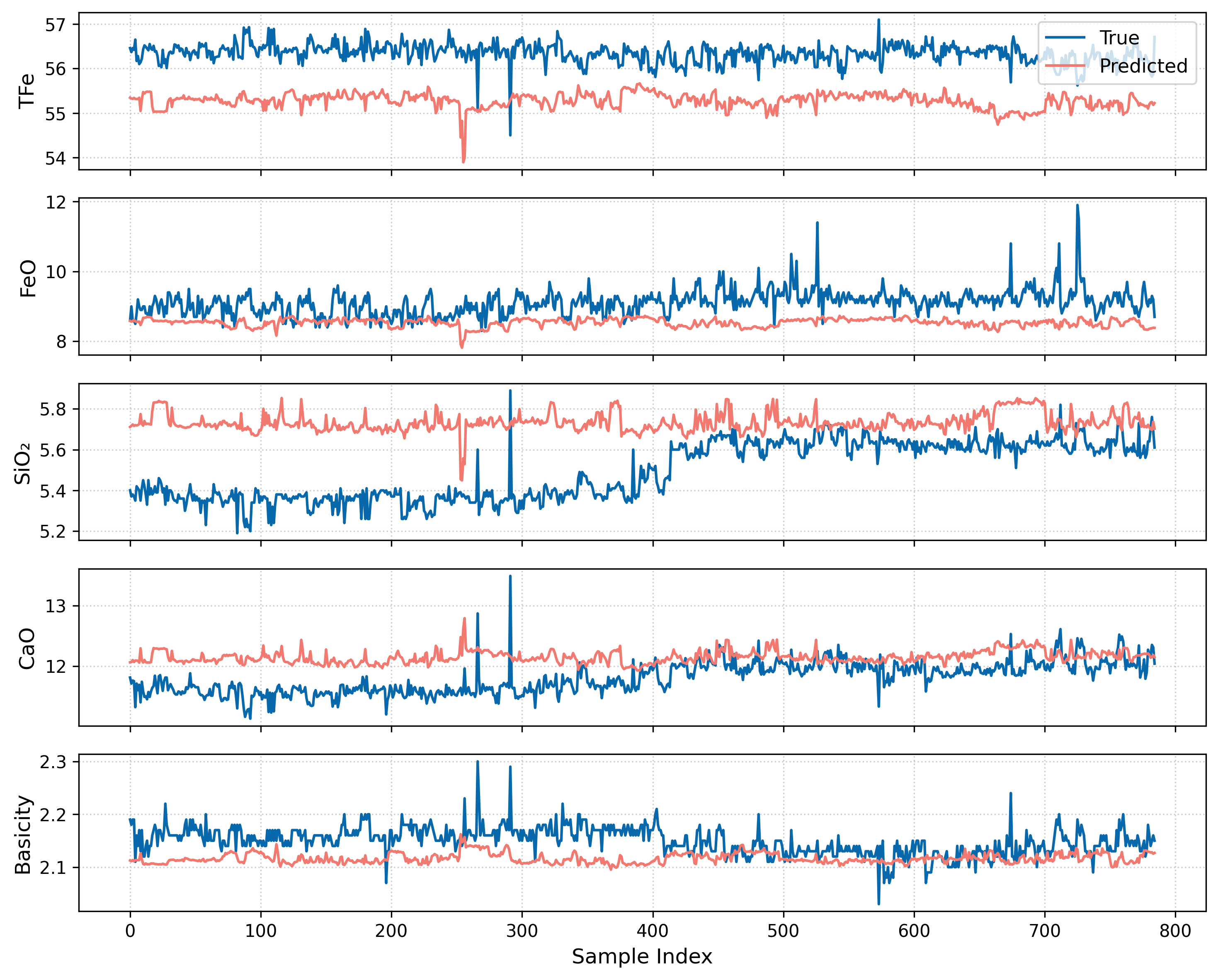}
  }

 \caption{Qualitative comparison of ground-truth and predicted trajectories for five quality targets under the chronological (non-shuffled) evaluation setting. Each panel shows the test-set predictions of one model (TFe, FeO, SiO\textsubscript{2}, CaO, and Basicity).}
  \label{fig:static_true_pred}
\end{figure}

\subsection{Ablation Study}
\label{sec:ablation}

\begin{table}[H]
\centering
\footnotesize
\setlength{\tabcolsep}{3.8pt}
\renewcommand{\arraystretch}{1.05}
\caption{PHMOEA ablation on H-DTLZ2 and H-DTLZ7 (mean$\pm$std across random seeds).}
\label{tab:abl_phmoea_app}
\begin{tabular}{@{}l l cc@{}}
\toprule
Benchmark & Variant & IGD$\downarrow$ & HV$\uparrow$ \\
\midrule
\multirow{3}{*}{H-DTLZ2}
& Full (PHMOEA) & 0.0051723$\pm$0.00008 & 0.4188054$\pm$0.00024 \\
& Without De-duplication & 0.0054982$\pm$0.00017 & 0.41825782$\pm$0.00029 \\
& Without Elitism & 0.0088575$\pm$0.00190 & 0.4135749$\pm$0.00629 \\
\midrule
\multirow{3}{*}{H-DTLZ7}
& Full (PHMOEA) & 0.0033303$\pm$0.00013 & 0.5566297$\pm$0.00001 \\
& Without De-duplication & 0.0033351$\pm$0.00016 & 0.5466567$\pm$0.00002 \\
& Without Elitism & 0.0102604$\pm$0.00013 & 0.54902670$\pm$0.00001 \\
\bottomrule
\end{tabular}

\end{table}

\begin{figure}[H]
  \centering
  \includegraphics[width=\linewidth]{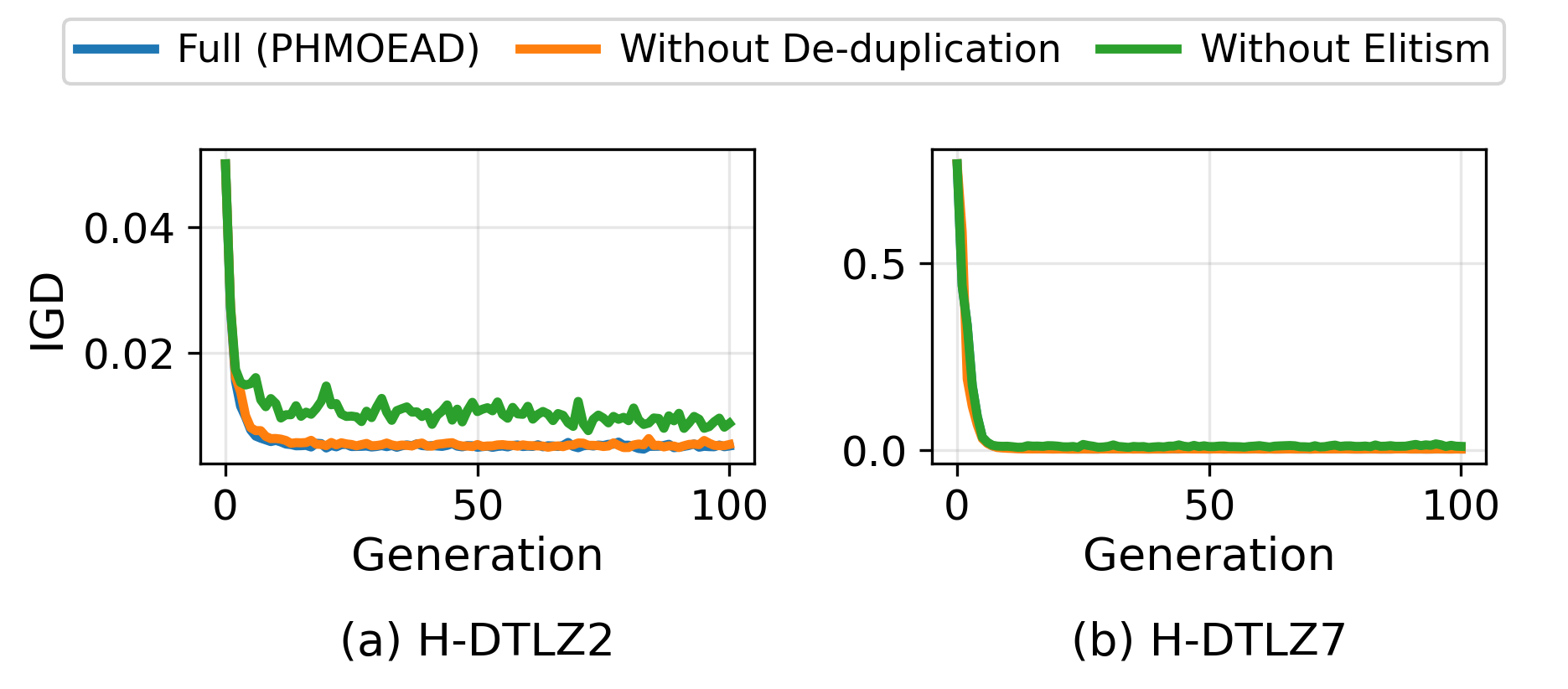}
  \caption{IGD trajectories over generations for PHMOEA ablations on synthetic benchmarks.}
  \label{fig:app_igd_curves}
\end{figure}

\begin{table}[H]
  \centering
  \caption{Per-target ablation results of MS--BCNN on the real-world sintering task under the chronological (non-shuffled) evaluation setting (mean$\pm$std across 5 random seeds).}
  \label{tab:ablation_detail}

  \setlength{\tabcolsep}{4pt}
  \renewcommand{\arraystretch}{1.05}

  \begin{tabular}{>{\raggedright\arraybackslash}p{2.6cm} l c c c}
    \toprule
    Variant & Target & MSE / NMSE$\downarrow$ & MAE / NMAE$\downarrow$ & MAPE(\%)$\downarrow$ \\
    \midrule

    \multirow{6}{2.6cm}{\raggedright\arraybackslash Full (MS-BCNN)}
      & TFe
      & \cellcolor{gray!50}{$0.387834 \pm 0.141648$}
      & \cellcolor{gray!50}{$0.529125 \pm 0.106408$}
      & \cellcolor{gray!50}{$0.939281 \pm 0.188755$}
       \\
      & FeO
      & $0.205710 \pm 0.046634$
      & $0.344918 \pm 0.047771$
      & $3.767097 \pm 0.507709$
       \\
      & SiO$_2$
      & $0.092268 \pm 0.037270$
      & $0.256975 \pm 0.064096$
      & $4.688675 \pm 1.153858$
       \\
      & CaO
      & $0.340601 \pm 0.136297$
      & $0.479478 \pm 0.110109$
      & $4.048819 \pm 0.919742$
       \\
      & Basicity
      & $0.001920 \pm 0.001217$
      & $0.035376 \pm 0.011207$
      & $1.639518 \pm 0.517844$
       \\
      & Overall
      & \cellcolor{gray!50}{$4.445824 \pm 2.193273$}
      & \cellcolor{gray!50}{$1.682359 \pm 0.490118$}
      & $3.016678 \pm 0.657582$
       \\
    \midrule

    \multirow{6}{2.6cm}{\raggedright\arraybackslash w/o Time Embedding}
      & TFe
      & $0.528306 \pm 0.438353$
      & $0.589665 \pm 0.291728$
      & $1.046420 \pm 0.517767$
       \\
      & FeO
      & \cellcolor{gray!50}{$0.196099 \pm 0.051730$}
      & \cellcolor{gray!50}{$0.339057 \pm 0.054657$}
      & \cellcolor{gray!50}{$3.741805 \pm 0.623491$}
       \\
      & SiO$_2$
      & $0.109426 \pm 0.063735$
      & $0.283855 \pm 0.096255$
      & $5.205599 \pm 1.756073$
       \\
      & CaO
      & \cellcolor{gray!50}{$0.284755 \pm 0.185126$}
      & $0.440510 \pm 0.163061$
      & \cellcolor{gray!50}{$3.735470 \pm 1.396127$}
       \\
      & Basicity
      & $0.004088 \pm 0.001677$
      & $0.056617 \pm 0.014191$
      & $2.622988 \pm 0.659113$
       \\
      & Overall
      & $5.660484 \pm 3.074876$
      & $1.905506 \pm 0.575237$
      & $3.270456 \pm 0.657582$
       \\
    \midrule

    \multirow{6}{2.6cm}{\raggedright\arraybackslash Single-branch (Short-only)}
      & TFe
      & $0.408283 \pm 0.224539$
      & $0.533293 \pm 0.204381$
      & $0.857912 \pm 0.362429$
       \\
      & FeO
      & $0.210917 \pm 0.049795$
      & $0.340264 \pm 0.050435$
      & $3.737282 \pm 0.523007$
       \\
      & SiO$_2$
      & \cellcolor{gray!50}{$0.085649 \pm 0.047021$}
      & $0.266748 \pm 0.077165$
      & $4.508693 \pm 1.423346$
       \\
      & CaO
      & $0.366627 \pm 0.065350$
      & $0.483729 \pm 0.065157$
      & $3.662865 \pm 0.569832$
       \\
      & Basicity
      & $0.003152 \pm 0.000510$
      & $0.048463 \pm 0.004712$
      & $2.241636 \pm 0.218776$
       \\
      & Overall
      & $4.910173 \pm 2.186641$
      & $1.801127 \pm 0.476171$
      & \cellcolor{gray!50}{$3.001678 \pm 0.657582$}
       \\
    \midrule

    \multirow{6}{2.6cm}{\raggedright\arraybackslash Single-branch (Long-only)}
      & TFe
      & $0.406314 \pm 0.053484$
      & $0.536952 \pm 0.058610$
      & $0.953975 \pm 0.104273$
       \\
      & FeO
      & $0.159779 \pm 0.013265$
      & \cellcolor{gray!50}{$0.296559 \pm 0.017777$}
      & $3.238290 \pm 0.190778$
       \\
      & SiO$_2$
      & $0.097965 \pm 0.037240$
      & $0.262124 \pm 0.054341$
      & $4.733682 \pm 0.986584$
       \\
      & CaO
      & $0.502150 \pm 0.165891$
      & $0.596165 \pm 0.134105$
      & $5.011213 \pm 1.122812$
       \\
      & Basicity
      & \cellcolor{gray!50}{$0.001816 \pm 0.000817$}
      & \cellcolor{gray!50}{$0.034605 \pm 0.008354$}
      & $1.601567 \pm 0.386090$
       \\
      & Overall
      & $4.956630 \pm 2.725779$
      & $1.753690 \pm 0.603572$
      & $3.107745 \pm 0.657582$
       \\
    \bottomrule
  \end{tabular}

  \vspace{1mm}
  {\scriptsize\textit{Note:} For each target row, we report MSE/MAE (with MAPE). The \textit{Overall} row reports NMSE and NMAE (NMSE = MSE/var, NMAE = MAE/std) macro-averaged over the five targets; MAPE is also macro-averaged.}
\end{table}

\end{document}